\documentclass{article} 

\usepackage{iclr2023_conference,times}

\usepackage{amsmath,amsfonts,bm}

\def\eqref#1{equation~\ref{#1}}

\def\1{\bm{1}}

\DeclareMathAlphabet{\mathsfit}{\encodingdefault}{\sfdefault}{m}{sl}
\SetMathAlphabet{\mathsfit}{bold}{\encodingdefault}{\sfdefault}{bx}{n}

\usepackage{hyperref}
\usepackage{url}
\usepackage{amsmath,amssymb,graphicx,color,algorithm, algpseudocode,booktabs,bm,relsize,enumitem,multirow,amsthm,epsfig,caption}

\usepackage[utf8]{inputenc} 
\usepackage[T1]{fontenc}    
\usepackage{booktabs}       
\usepackage{amsfonts}       
\usepackage{nicefrac}       
\usepackage{microtype}      
\usepackage{xcolor}         

\usepackage{xspace}
\usepackage{listings}
\usepackage{subcaption}
\usepackage{wrapfig}
\usepackage{lipsum}
\usepackage{enumitem}
\usepackage{chngpage}

\newcommand{\Algname}{\textbf{P}lanning-guided self-\textbf{I}mitation learning for \textbf{G}oal-conditioned policies\xspace}

\newcommand{\ALGname}{PIG\xspace}
\newcommand{\LthreeP}{$L^{3}P$\xspace}

\definecolor{Blue9}{rgb}{0.098,0.3,0.9}
\definecolor{Red7}{rgb}{0.941, 0.243, 0.243}
\definecolor{Green7}{RGB}{55, 178, 77}
\definecolor{BrickRed}{rgb}{0.6,0,0}
\definecolor{RoyalBlue}{rgb}{0,0,0.8}
\definecolor{Tdgreen}{rgb}{0,0.4,0.7}
\definecolor{JSViolet}{RGB}{71,15,244}
\definecolor{JSRed}{RGB}{205,44,78}
\definecolor{cadmiumgreen}{rgb}{0.0, 0.42, 0.24}
\definecolor{Teal4}{RGB}{56, 217, 169}
\definecolor{Cyan4}{RGB}{59, 201, 219}

\hypersetup{colorlinks=true}
\hypersetup{citecolor=JSViolet, linkcolor=JSRed}

\newcommand{\highlight}[1]{{\color{JSViolet} #1}}

\title{Imitating Graph-Based Planning with \\ Goal-Conditioned Policies}

\iclrfinalcopy

\author{Junsu Kim$^{1}$, Younggyo Seo$^{1}$, Sungsoo Ahn$^{2}$, Kyunghwan Son$^{1}$, Jinwoo Shin$^{1}$ \\
$^1$ Korea Advanced Institute of Science and Technology (KAIST) \\
$^2$ Pohang University of Science and Technology (POSTECH) \\
\texttt{\{junsu.kim, younggyo.seo, kevinson9473, jinwoos\}@kaist.ac.kr} \\
\texttt{sungsoo.ahn@postech.ac.kr}
}

\begin{document}

\maketitle

\begin{abstract}
Recently, graph-based planning algorithms have gained much attention to solve goal-conditioned reinforcement learning (RL) tasks:
they provide a sequence of subgoals to reach the target-goal, and the agents learn to execute subgoal-conditioned policies. However, the sample-efficiency of such RL schemes still remains a challenge, particularly for long-horizon tasks. 
To address this issue, we present a simple yet effective self-imitation
scheme which distills a subgoal-conditioned policy into the target-goal-conditioned policy.
Our intuition here is that to reach a target-goal, an agent should pass through a subgoal, so target-goal- and subgoal- conditioned policies should be similar to each other.
We also propose a novel scheme of stochastically skipping executed subgoals in a planned path, which further improves performance. 
Unlike prior methods that only utilize graph-based planning in an execution phase, our method transfers knowledge from a planner along with a graph into policy learning. 
We empirically show that our method can significantly boost the sample-efficiency of the existing goal-conditioned RL methods under various long-horizon control tasks.\footnote{Code is available at \url{https://github.com/junsu-kim97/PIG}}
\end{abstract}

\section{Introduction}
Many sequential decision making problems can be expressed as reaching a given goal, e.g., navigating a walking robot 
\citep{schaul2015universal,nachum2018data} and fetching an object using a robot arm \citep{andrychowicz2017hindsight}. 
Goal-conditioned reinforcement learning (GCRL) aims to solve this problem by training a goal-conditioned policy to guide an agent towards reaching the target-goal. 
In contrast to many of other reinforcement learning frameworks, GCRL is capable of solving different problems (i.e., different goals) using a single policy.

\begin{figure}[t!]
\vspace{-.2in}
\centering
\begin{subfigure}{0.325\textwidth}
\centering
\includegraphics[width=1.0\textwidth]{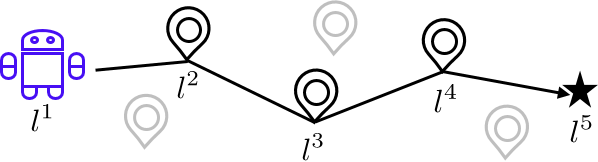}
\caption{
If $(l^{1}, l^{2}, l^{3}, l^{4}, l^{5})$ is an optimal $l^{5}$-reaching path, all the sub-paths are optimal for reaching $l^{5}$.
}\label{fig:comp1}
\end{subfigure}
\vspace{10mm}
\hfill
\begin{subfigure}{0.66\textwidth}
\centering
\includegraphics[width=1.0\textwidth]{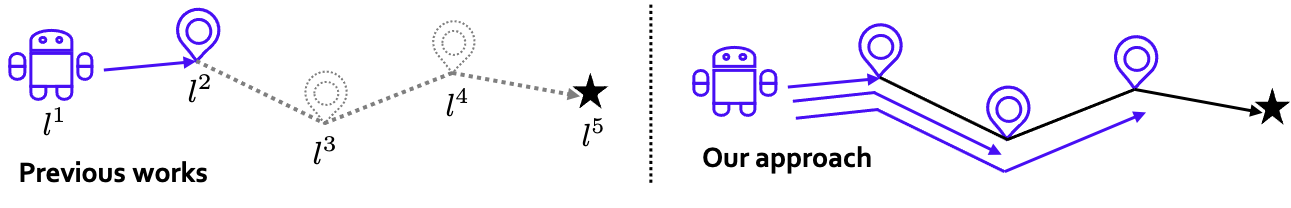}
\caption{
Previous works guide the agent using a $l^{2}$-reaching sub-path. Our work uses all the possible sub-paths that reach $l^{2}, l^{3}, l^{4}, l^{5}$.
}\label{fig:comp2}
\end{subfigure}
\vspace{-10mm}
\caption{Illustration of (a) optimal substructure property
and (b) sub-paths considered in previous works and our approach for guiding the training of a goal-reaching agent.
}
\label{fig:comp}
\end{figure}

An intriguing characteristic of GCRL is its \textit{optimal substructure property}; any sub-path of an optimal goal-reaching path is an optimal path for its endpoint (Figure~\ref{fig:comp1}). 
This implies that a goal-conditioned policy is replaceable by a policy conditioned on a ``subgoal'' existing between the goal and the agent. 
Based on this insight, researchers have investigated graph-based planning to construct a goal-reaching path by (a) proposing a series of subgoals and (b) 
executing policies conditioned on the nearest subgoal \citep{savinov2018semiparametric,eysenbach2019search, huang2019mapping}. 
Since the nearby subgoals are easier to reach than the faraway goal, such 
planning improves the success ratio of the agent reaching the target-goal during sample collection. 

In this paper, we aim to improve the existing GCRL algorithms to be even more faithful to the optimal substructure property.
To be specific, we first incorporate the optimal substructure property into the training objective of GCRL to improve the sample collection algorithm.
Next, when executing a policy, we consider using all the proposed subgoals as an endpoint of sub-paths instead of 
using just the subgoal nearest to the agent (Figure~\ref{fig:comp2}). 

\textbf{Contribution.}
We present \Algname (\ALGname), a novel and generic framework that builds upon the existing GCRL frameworks 
that use
graph-based planning.
\ALGname consists of the following key ingredients (see Figure~\ref{fig:concept}):
\begin{itemize}[topsep=1.0pt,itemsep=1.0pt,leftmargin=5.5mm]
    \item [$\bullet$] \textbf{Training with self-imitation:} 
    we propose a new training objective that encourages a goal-conditioned policy to imitate the subgoal-conditioned policy. Our intuition is that policies conditioned on nearby subgoals are more likely to be accurate than the policies conditioned on a faraway goal.
    In particular, we consider the imitation of policies conditioned on all the subgoals proposed by the graph-based planning algorithm. 
    \item [$\bullet$] \textbf{Execution\footnote{In this paper, we use the term ``execution'' to denote both (1) the roll-out in training phase and (2) the deployment in test phase.} with subgoal skipping:}
    As an additional technique that fits our self-imitation loss,
    we also propose \textit{subgoal skipping}, which randomizes a subgoal proposed by the graph-based planning to further improve the sample-efficiency.
    During the sample-collection stage and deployment stage, policies randomly ``skip'' conditioning on some of the subgoals proposed by the planner when it is likely that the learned policies can reach the proposed subgoals. Such a procedure is based on our intuition that an agent may find a better goal-reaching path by ignoring some subgoals proposed by the planner when the policy is sufficiently trained with our loss.
\end{itemize}

We demonstrate the effectiveness of \ALGname on various long-horizon continuous control tasks based on MuJoCo simulator \citep{todorov2012mujoco}. In our experiments, \ALGname significantly boosts the sample-efficiency of an existing GCRL method, i.e., mapping state space 
(MSS) \citep{huang2019mapping},\footnote{{We note that \ALGname is a generic framework that can be also incorporated into any planning-based GCRL methods, other than MSS.
Nevertheless, we choose MSS because it is one of the most representative GCRL works 
as most recent works \citep{hoang2021successor, zhang2021world} could be considered as variants of MSS.
}} particularly in long-horizon tasks. For example, MSS + \ALGname achieves the success rate of 57.41\% in Large U-shaped AntMaze environment, while MSS only achieves 19.08\%. 
Intriguingly, we also find that the \ALGname-trained policy 
performs competitively 
even without any planner; 
this could be useful in some real-world scenarios where planning cost (time or memory) is expensive \citep{bency2019neural, qureshi2019motion}.
\section{Related work}

\textbf{Goal-conditioned reinforcement learning (GCRL).} 
GCRL aims to solve multiple tasks associated with target-goals \citep{andrychowicz2017hindsight, kaelbling1993learning, schaul2015universal}.
Typically, GCRL algorithms rely on the  universal value function approximator
(UVFA) \citep{schaul2015universal}, which is a single neural network that estimates the true value function given not just the states but also the target-goal. 
Furthermore, researchers have also investigated goal-exploring algorithms \citep{mendonca2021discovering, pong2020skew} to avoid any local optima of training the goal-conditioned policy. 

\textbf{Graph-based planning for GCRL.}
To solve long-horizon GCRL problems, graph-based planning can guide the agent to condition its policy on a series of subgoals that are easier to reach than the faraway target goal \citep{eysenbach2019search, hoang2021successor, huang2019mapping, laskin2020sparse, savinov2018semiparametric, zhang2021world}.
To be specific, the corresponding frameworks build a graph where nodes and edges correspond to states and inter-state distances, respectively. Given a shortest path between two nodes representing the current state and the target-goal, the policy conditions on a subgoal represented by a subsequent node in the path. 

For applying graph-based planning to complex environments, recent progress has mainly been made in building a graph that represents visited state space well while being scalable to large environments.
For example, \citet{huang2019mapping} and \citet{hoang2021successor} limits the number of nodes in a graph and makes nodes to cover visited state space enough by containing nodes that are far from each other in terms of L2 distance or successor feature similarity, respectively. Moreover, graph sparsification via two-way consistency \citep{laskin2020sparse} or learning latent space with temporal reachability and clustering \citep{zhang2021world} also have been proposed.
They have employed graph-based planning for providing the nearest subgoal to a policy at execution time, which utilizes the optimal substructure property in a limited context. In contrast, PIG aims to faithfully utilize the property both in training and execution via self-imitation and subgoal skipping, respectively.

\textbf{Self-imitation learning for goal conditioned policies.}
Self-imitation learning strengthens the training signal by imitating trajectories sampled by itself \citep{oh2018self,ding2019goal,chane2021goal,ghosh2021learning}. 
GoalGAIL \citep{ding2019goal} imitates goal-conditioned actions from expert demonstrations along with the goal-relabeling strategy \citep{andrychowicz2017hindsight}. 
Goal-conditioned supervised learning 
(GCSL) \citep{ghosh2021learning}
trains goal-conditioned policies via iterated supervised learning with goal-relabeling.
RIS \citep{chane2021goal} makes target-goal- and subgoal- conditioned policy be similar, where the subgoal is from a high-level policy that is jointly trained with a (low-level) policy. Compared to prior works, \ALGname faithfully incorporates optimal substructure property with two distinct aspects: (a) graph-based planning and (b) actions from a current policy rather than past actions, where we empirically find that these two differences are important for performance boost (see Section~\ref{sec:bc_loss}). Nevertheless, we remark that \ALGname is an orthogonal framework to them, so applying \ALGname on top of them (e.g., RIS) would be an interesting future work (e.g., leveraging both planning and high-level policy).

\textbf{Distilling planning into a policy.}
Our idea of distilling outcomes of planner into the goal-conditioned policy is connected to prior works in the broader planning context. For example, AlphaGo Zero \citep{silver2017mastering} distills the outcome of the Monte-Carlo Tree Search (MCTS) planning procedure into a prior policy. Similarly, SAVE \citep{Hamrick2020Combining} distills the MCTS outcomes into the action-value function. \ALGname aligns with them in that we distill planned-subgoal-conditioned policy into the target-goal-conditioned policy.
\section{Preliminary: goal-conditioned RL with graph-based planning}
\begin{figure*}[t]
    \vspace{-.2in}
    \centering
    \includegraphics[width=1.0\textwidth]{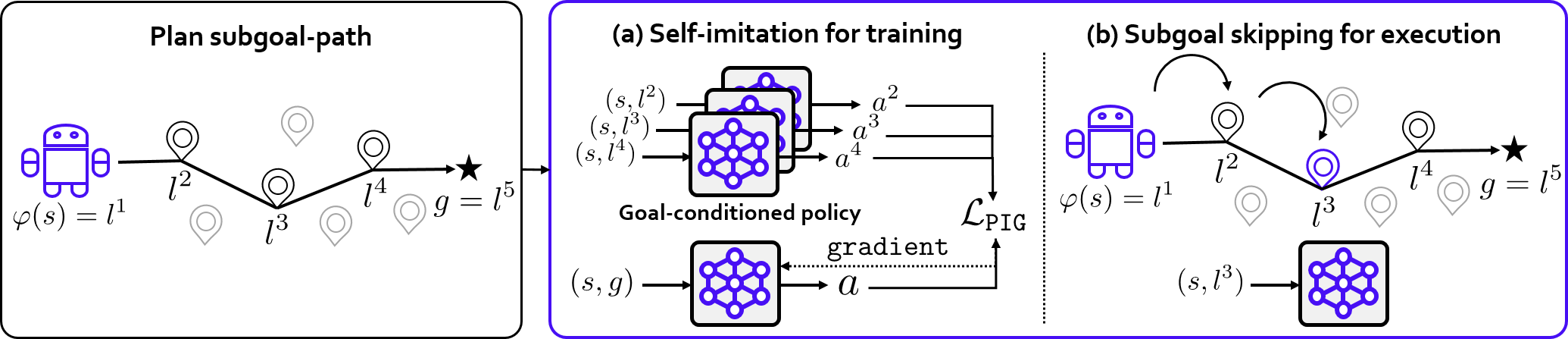}
    \caption{Illustration of \Algname (\ALGname). The key ingredient of \ALGname is twofold: (a) self-imitation for training and (b) subgoal skipping for execution. For (a), we distill a planned-subgoal-conditioned policy into the target-goal-conditioned policy via our self-imitation loss term $\mathcal{L}_{\mathtt{\ALGname}}$. A policy is trained using the auxiliary $\mathcal{L}_{\mathtt{\ALGname}}$ along with off-the-shelf actor loss. 
    For (b), we randomize a subgoal provision from a planner.
    }
    \label{fig:concept}
\end{figure*}
In this section, we describe the existing graph-based planning framework for goal-conditioned reinforcement learning, upon which we build our work.
To this end, in Section~\ref{sec:gcrl}, we describe the problem setting of GCRL. Next, in Section~\ref{sec:training}, we explain how to train the goal-conditioned policy using hindsight experience replay \citep{andrychowicz2017hindsight}. Finally, in Section~\ref{sec:planning}, we explain how graph-based planning can help the agent to execute better policy. We provide the overall pipeline in 
Algorithm~\ref{alg:framework} in Supplemental material~\ref{supp:algo_table}, colored as black.

\subsection{Problem description}
\label{sec:gcrl}

We formulate our control task as a finite-horizon, goal-conditioned Markov decision process 
(MDP) \citep{sutton2018reinforcement}
as a tuple $(\mathcal{S}, \mathcal{G}, \mathcal{A}, p, r, \gamma, H)$ corresponding to state space $\mathcal{S}$, goal space $\mathcal{G}$, action space $\mathcal{A}$, transition dynamics  $p\left(s^\prime|s,a\right)$ for $s, s^{\prime} \in \mathcal{S}, a \in \mathcal{A}$, reward function $r\left(s, a, s^{\prime}, g\right)$, discount factor $\gamma \in [0,1)$, and horizon $H$. 

Following prior works \citep{huang2019mapping, zhang2021world}, we consider a setup where every state can be mapped into the goal space using a goal mapping function $\varphi:\mathcal{S}\rightarrow \mathcal{G}$. Then the agent attempts to reach a certain state $s$ associated with the target-goal $g$, i.e., $\varphi(s)=g$. For example, for a maze-escaping game with continuous locomotion, each state $s$ represents the location and velocity of the agent, while the goal $g$ indicates a certain location desired to be reached by the agent.

Typically, GCRL considers the reward function defined as follows:
\begin{align}
\begin{split}
\label{eq:reward}
    r(s,a, s^{\prime}, g) = \begin{cases}
                    0 \quad & \Vert \varphi(s') - g \Vert_{2} \leq \delta \\
                    -1 \quad & \text{otherwise}
               \end{cases}
\end{split}
\end{align}
where $\delta$ is a pre-set threshold to determine whether the state $s'$ from the transition dynamics $p(s^\prime|s, a)$ is close enough to the goal $g$. 
To solve GCRL, we optimize a deterministic goal-conditioned policy $\pi : \mathcal{S} \times \mathcal{G} \rightarrow \mathcal{A}$ to maximize the expected cumulative future return $V_{g, \pi} \left(s_{0}\right) = \sum_{t=0}^{\infty} \gamma^{t} r(s_{t}, a_{t}, s_{t+1}, g)$ where $t$ denotes timestep and $a_{t} = \pi(s_{t}, g)$.

\subsection{Training with hindsight experience replay}
\label{sec:training}
To train goal-conditioned policies, any off-the-shelf RL algorithm can be used.
Following prior works \citep{huang2019mapping, zhang2021world}, we use deep deterministic policy gradient 
(DDPG) \citep{Lillicrap2015continuous}
as our base RL algorithm.
Specifically, we train an action-value function (critic) $Q$ with parameters $\phi$ and a deterministic policy (actor) $\pi$ with parameters $\theta$ given a replay buffer $\mathcal{B}$, by optimizing the following losses:
\begin{align}
\begin{split}
\label{eq:ddpg_critic}
    \mathcal{L}_{\mathtt{critic}} (\phi) = \mathbb{E}_{(s_{t}, a_{t}, r_{t}, g) \sim \mathcal{B}} \bigg[ (Q_{\phi}(s_{t}, a_{t}, g) - y_{t})^{2} \bigg] \\
    \text{where}\; y_{t} = r_{t} + \gamma Q_{\phi}(s_{t+1}, \pi_{\theta}(s_{t+1}, g), g) 
\end{split}
\end{align}
\begin{align}
\label{eq:ddpg_actor}
    \mathcal{L}_{\mathtt{actor}} (\theta) = - \mathbb{E}_{(s_{t}, a_{t}, g) \sim \mathcal{B}} [Q_{\phi}(s_{t}, \pi_{\theta}(s_{t}, g), g) ],
\end{align}
where the critic $Q_{\phi}$ is a universal value function approximator 
(UVFA) \citep{schaul2015universal}
trained to estimate the goal-conditioned action-value.
However, it is often difficult to train UVFA because the target-goal can be far from the initial position, which makes the agents unable to receive any reward signal.
To address the issue, goal-relabeling technique proposed in hindsight experience replay 
(HER) \citep{andrychowicz2017hindsight}
is widely-used for GCRL methods.
The key idea of HER is to reuse any trajectory ending with state $s$ as supervision for reaching the goal $\varphi(s)$. This allows for relabelling any trajectory as success at hindsight even if the agent failed to reach the target-goal during execution.

\subsection{Execution with graph-based planning}
\label{sec:planning}
In prior works \citep{huang2019mapping, zhang2021world},
graph-based planning provides a subgoal, which is a waypoint to reach a target goal when executing a policy. 
A planner runs on 
a weighted graph that abstracts visited state space.

\textbf{Graph construction.}
The planning algorithms build a weighted directed graph $\mathcal{H}=(\mathcal{V}, \mathcal{E}, d)$ where each node $l \in \mathcal{V} \subseteq \mathcal{G}$ is specified by a state $s$ visited by the agent, i.e., $l=\varphi(s)$. 
For populating states, we execute the two-step process following \citet{huang2019mapping}: (a) random sampling of a fixed-sized pool from an experience replay and (b) farthest point sampling \citep{vassilvitskii2006k} from the pool to build the final collection of landmark states.
Then each edge 
$(l^{1}, l^{2}) \in \mathcal{E}$
is assigned for any pair of states that can be visited from one to another by a single transition in the graph. 
A weight $d(l^{1}, l^{2})$ is an estimated distance between the two nodes, i.e., the minimum number of actions required for the agent to visit node $l^{2}$ starting from $l^{1}$. Given $\gamma \approx 1$ and the reward shaped as in Equation~\ref{eq:reward}, one can estimate the distance $d(l^{1}, l^{2})$ as the corresponding value function $-V(s^{1}, l^{2}) \approx -Q_{\phi}(s^{1}, a^{1,2}, l^{2})$ where $l^{2} = \varphi(s^{2})$ and $a^{1,2} = \pi_{\theta}(s^{1}, l^{2})$ \citep{huang2019mapping}.
Next, we link all the nodes in the graph and give a weight $d (\cdot, \cdot)$ for each generated edge. 
Then, if a weight of an edge is greater than (pre-defined) threshold, cut the edge.
We provide further details of graph construction in Supplemental material~\ref{supp:graph_construction}.

\textbf{Planning-guided execution.}
The graph-based planning provides a policy with an emergent node to visit when executing the policy. 
To be specific, given a graph $\mathcal{H}$, a state $s$ and, a target goal $g$, 
we expand the graph by appending $s$ and $g$, and obtain
a shortest path $\tau_{g} = (l^{1},\ldots, l^{N})$ such that $l^{1} = \varphi(s)$ and $l^{N}=g$ using a planning algorithm. Then, a policy is conditioned on a nearby subgoal $l^{2}$, which is easier to reach than the faraway target-goal $g$. 
This makes it easy for the agent to collect successful samples reaching the target goals, leading to an overall performance boost.
Note that we re-plan for every timestep following prior works \citep{huang2019mapping, zhang2021world}.
\section{Planning-guided self-imitation learning for GCRL}
In this section, we introduce a new framework, named \ALGname, for improving the sample-efficiency of GCRL using graph-based planning. Our framework adds two components on top of the existing methods: (a) training with self-imitation and (b) execution with subgoal skipping, which highlights the generality of our concept 
(colored as \highlight{purple} in Algorithm~\ref{alg:framework} in Supplementary material~\ref{supp:algo_table}).
Our main idea fully leverages the optimal substructure property; any sub-path of an optimal goal-reaching path is an optimal path for its endpoint (Figure~\ref{fig:comp1}).
 In the following sections, we explain our self-imitation loss as a new training objective in Section~\ref{subsec:loss} and subgoal skipping strategy for execution in Section~\ref{subsec:coarse}.
 We provide an illustration of our framework in Figure~\ref{fig:concept}.

\subsection{Training with self-imitation}
    \label{subsec:loss}
Motivated by the intuition that an agent should pass through a subgoal to reach a target-goal, we encourage actions from target-goal- and subgoal- conditioned policy to stay close, where the subgoals are nodes in a planned subgoal-path. By doing so, we expect that faraway goal-conditioned policy learns plausible actions that are produced by (closer) subgoal-conditioned policy. Specifically, we devise a loss term $\mathcal{L}_{\mathtt{\ALGname}}$ given a stored planned path $\tau_{g} = (l^{1}, l^{2}, \ldots, l^{N})$ and a transition $(s, g, \tau_{g})$ from a replay buffer $\mathcal{B}$ as follows:
\begin{align}
\label{eq:our_bc_loss}
\mathcal{L_{\mathtt{\ALGname}}} (\theta) = \mathbb{E}_{(s, \tau_{g}, g) \sim \mathcal{B}} \bigg[\frac{1}{N-1} \sum_{l^{k} \in \tau_{g} \setminus \{l^{1}\}} \Vert \pi_{\theta}(s, g) - \mathtt{SG} (\pi_{\theta}(s, l^{k})) \Vert_{2}^{2} \bigg]
\end{align}
where $\mathtt{SG}$ refers to a stop-gradient operation.
Namely, the goal-conditioned policy imitates behaviors of subgoal-conditioned policy. We incorporate our self-imitation loss term into the existing GCRL frameworks by plugging $\mathcal{L}_{\mathtt{\ALGname}}$ as an extra loss term into the original policy loss term as follows:
\begin{align}
\label{eq:total_loss}
\mathcal{L} (\theta) = \mathcal{L}_{\mathtt{actor}} (\theta) + \lambda \mathcal{L}_{\mathtt{\ALGname}} (\theta)
\end{align}
where $\lambda$ is a balancing coefficient, which is a pre-set hyperparameter.

One can also understand that self-imitating loss improves performance by enhancing the correctness of planning. Note that actor is used to estimate distance $d$ between two nodes $l^{1}, l^{2}$; $d(l^{1}, l^{2}) \approx -Q_{\phi}(s^{1}, \pi_{\theta} (s^{1}, l^{2}), l^{2})$ as mentioned in Section~\ref{sec:planning}. Our self-imitating loss makes $\pi_{\theta}$ more accurate for even faraway goals, so it leads to the precise construction of a graph. Then, planning gives more suitable subgoals for an actor in execution.

\subsection{Execution with subgoal skipping}
\label{subsec:coarse}
As an additional technique that fits our self-imitation loss, we propose \textit{subgoal skipping}, which randomizes a subgoal proposed by the graph-based planning to further improve the sample-efficiency.
Note that the existing graph-based planning for GCRL always provides the nearest node $l^{2}$ in the searched path $\tau_{g}$ as a desired goal $l^{*}$ regardless of how a policy is trained. 
Motivated by our intuition that an agent may find a better goal-reaching path (i.e., short-cuts) by ignoring some of the subgoals, we propose a new subgoal selecting strategy.

Our subgoal skipping is based on the following insight: 
when a policy for the planned subgoal and the final goal agree (small $\mathcal{L}_{\mathtt{\ALGname}}$), diversifying subgoal suggestions could help find unvisited routes. 
Namely, the goal-conditioned policy is likely to be trustworthy if final-goal- and planned-subgoal- conditioned policies align because it implies that the goal-conditioned policy have propagated information quite far. 
Leveraging generalization capability of the trained policy, suggesting the policy with diversified subgoals rather than only the nearest subgoal could help finding better routes.

To be specific, to select the desired goal $l^{*}$, we start from the nearest node $l^{2}$ in the planned shortest path $\tau_{g}$, and stochastically jump to the next node until our condition becomes unsatisfied with the following binomial probability:
\begin{align}
\begin{split}
\label{eq:path_hop}    
    P[\mathtt{jump}] = \min \bigg(\frac{\alpha}{\mathcal{L}_{\mathtt{\ALGname, latest}}}, 1 \bigg),
\end{split}
\end{align}
where $\alpha$ is pre-set skipping temperature and $\mathcal{L}_{\mathtt{\ALGname, latest}}$ denotes $\mathcal{L}_{\mathtt{\ALGname}}$ calculated at the latest parameter update. We set $l^{*}$ as the final subgoal after the jumping.
Intuitively, the jumping criterion is likely to jump more for a smaller $\mathcal{L}_{\mathtt{\ALGname, latest}}$, which is based on the fact that the policy is likely to behave more correctly for a faraway goal. 
As subgoals are sampled from the searched shortest path and it is not likely to choose a farther subgoal if a policy is not trustworthy for faraway subgoals, our sampled subgoals are likely to be appropriate for a current policy.
We describe our subgoal skipping procedure in Algorithm~\ref{alg:skip} of Supplemental material ~\ref{supp:algo_table}.

\section{Experiment}
\label{sec:experiment}
\begin{figure*}[t!]
    \vspace{-.3in}
    \centering
    \begin{subfigure}{0.115\textwidth}
    \includegraphics[width=1.0\linewidth]{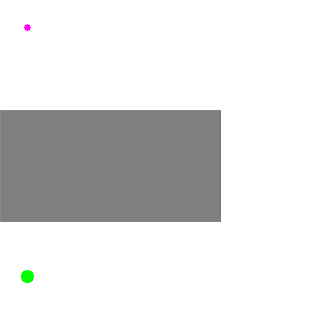}
    \caption{2DReach}
    \end{subfigure}
    \begin{subfigure}{0.115\textwidth}
    \includegraphics[width=1.0\linewidth]{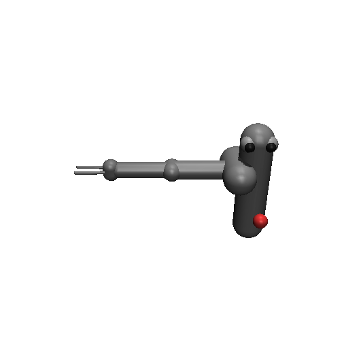}
    \caption{Reacher}
    \end{subfigure}
    \begin{subfigure}{0.115\textwidth}
    \includegraphics[width=1.0\linewidth]{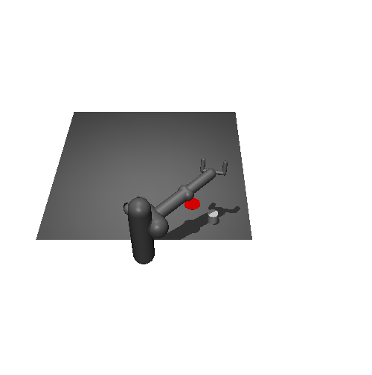}
    \caption{Pusher}
    \end{subfigure}
    \begin{subfigure}{0.115\textwidth}
    \includegraphics[width=1.0\linewidth]{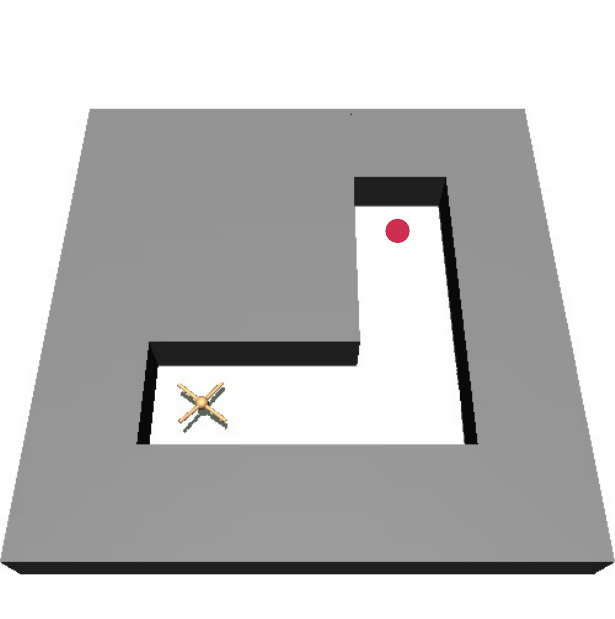}
    \caption{L-shape}
    \end{subfigure}
    \begin{subfigure}{0.115\textwidth}
    \includegraphics[width=1.0\linewidth]{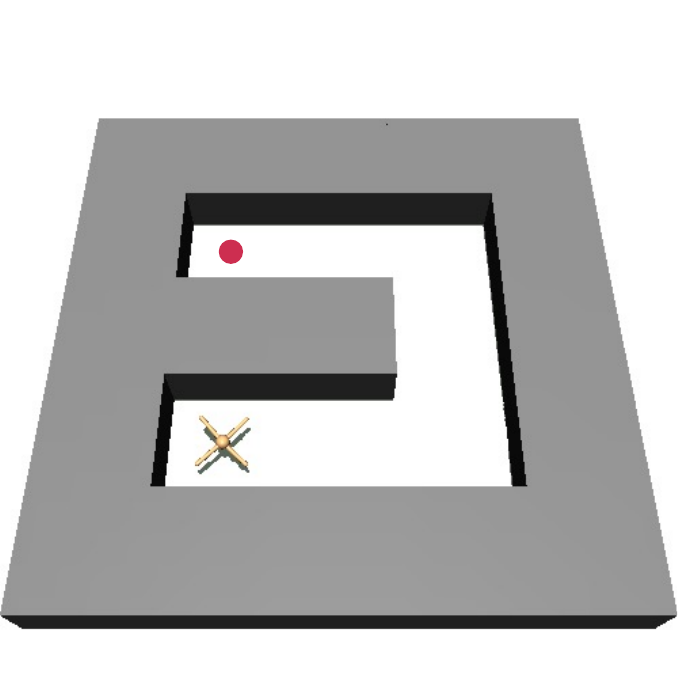}
    \caption{U-shape}
    \end{subfigure}
    \begin{subfigure}{0.115\textwidth}
    \includegraphics[width=1.0\linewidth]{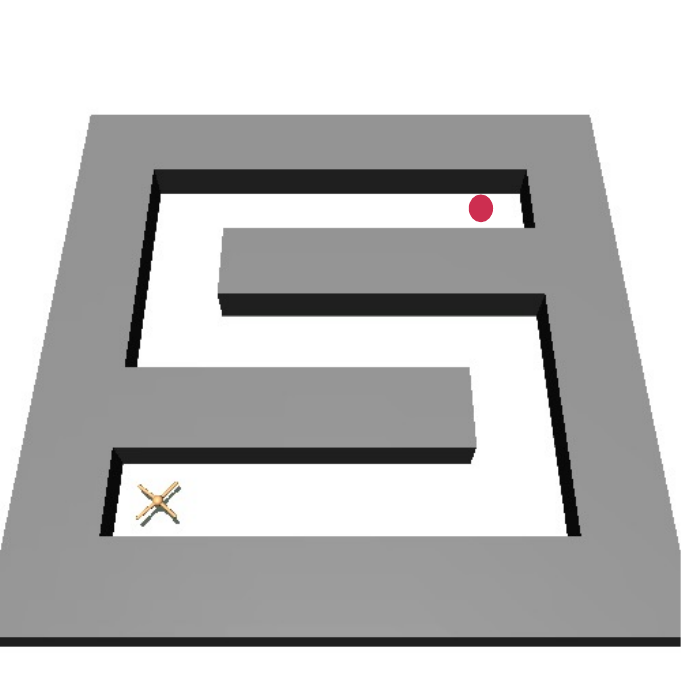}
    \caption{S-shape}
    \end{subfigure} 
    \begin{subfigure}{0.115\textwidth}
    \includegraphics[width=1.0\linewidth]{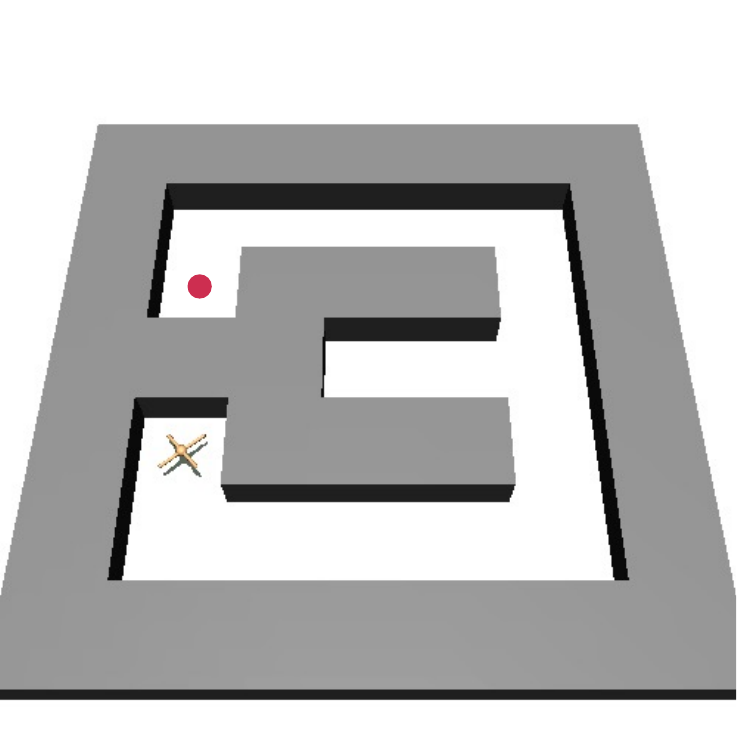}
    \caption{$\omega$-shape}
    \end{subfigure}
    \begin{subfigure}{0.115\textwidth}
    \includegraphics[width=1.0\linewidth]{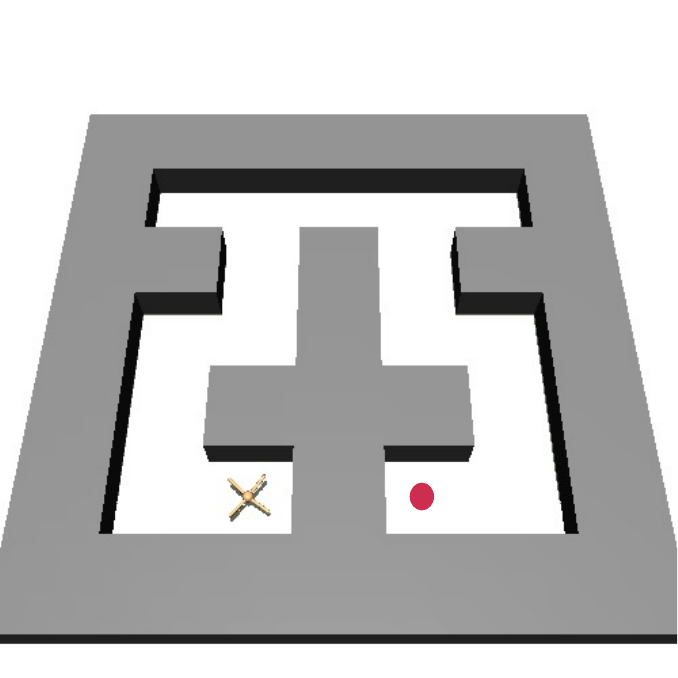}
    \caption{$\Pi$-shape}
    \end{subfigure}
    \caption{Environments used in our experiments. In all environments, at training time, an agent starts at a random point, and aims to reach a target goal that is set randomly. At test time for AntMaze tasks, the red point and the position of an ant indicates the target goal, and the initial point, respectively.}
    \label{fig:environment}
\end{figure*}
In this section, we design our experiments to answer the following questions:
\begin{itemize}[topsep=1.0pt,itemsep=1.0pt,leftmargin=5.5mm]
    \setlength\itemsep{-.05mm}
    \item Can \ALGname improve the sample-efficiency on long-horizon continuous control tasks over baselines (Figure~\ref{fig:main})?
    \item Can a policy trained by \ALGname perform well even without a planner at the test time (Figure~\ref{fig:planner})?
    \item How does \ALGname compare to another self-imitation strategy (Figure~\ref{fig:ablation})?
    \item Is the subgoal skipping effective for sample-efficiency (Figure~\ref{fig:skipping})?
    \item How does the balancing coefficient $\lambda$ affect performance (Figure~\ref{fig:lambda})?
\end{itemize}
\vspace{-.05in}
\subsection{Experimental setup}
\textbf{Environments.}
\begin{figure*}[t]
\vspace{-.2in}
\centering
\begin{subfigure}[t]{0.99\textwidth}
\centering
  \includegraphics[width=0.55\linewidth]{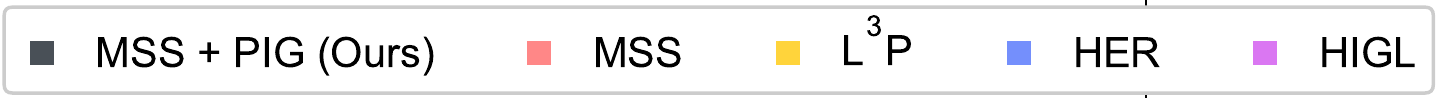}
  \vspace{0.05in}
  \label{fig:Legend1}
\end{subfigure}
\hspace*{\fill}
\begin{subfigure}{0.325\textwidth}
\includegraphics[width=1.0\linewidth]{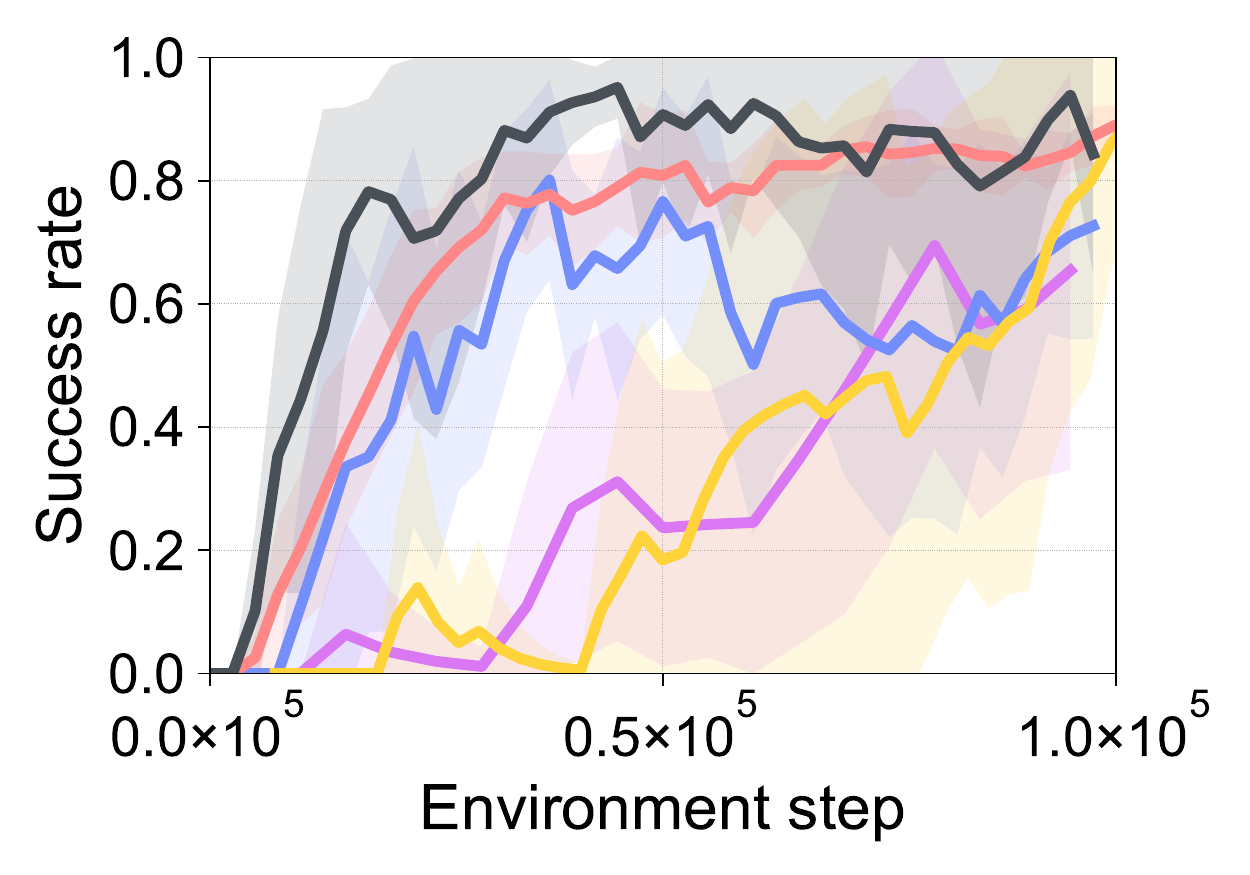}
\caption{2DReach}
\end{subfigure}
\begin{subfigure}{0.325\textwidth}
\includegraphics[width=1.0\linewidth]{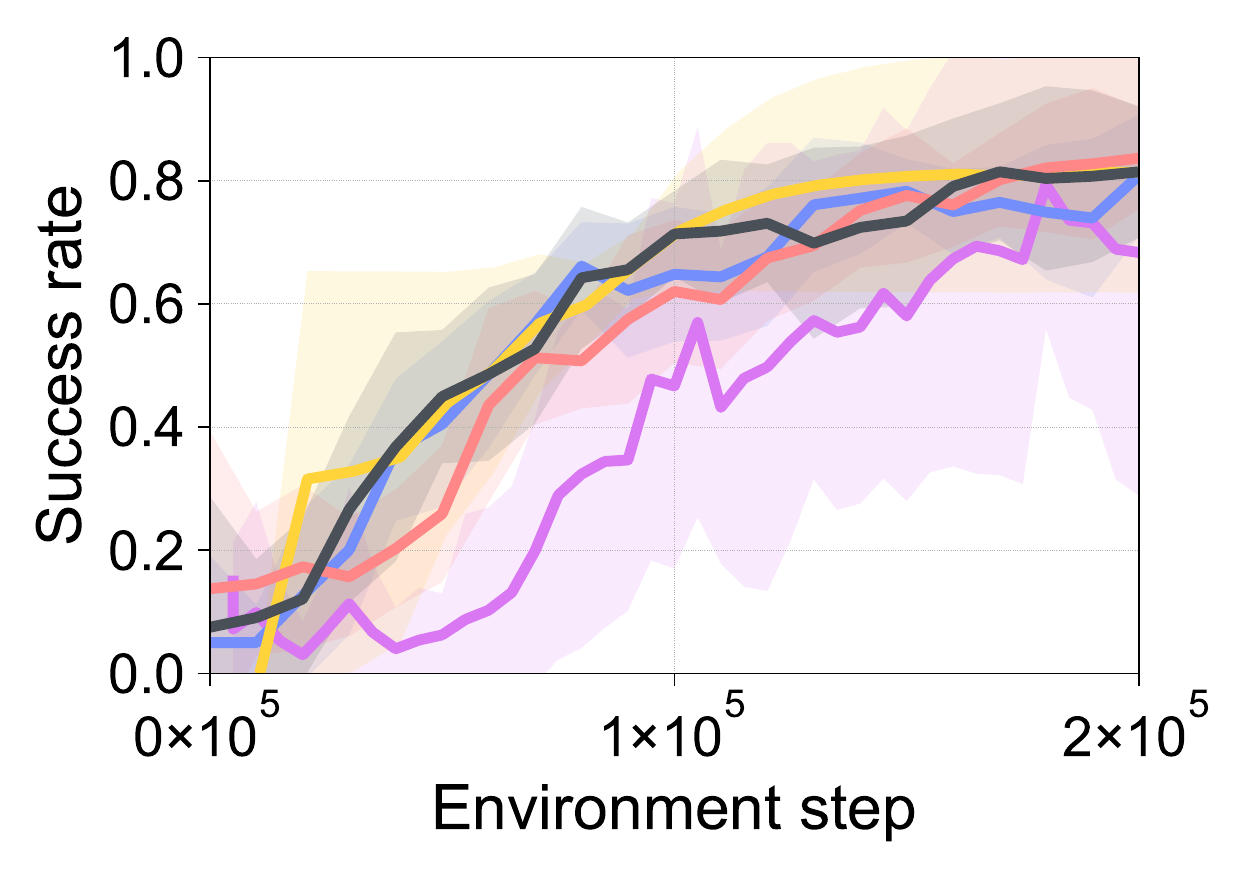}
\caption{Reacher}
\end{subfigure}
\begin{subfigure}{0.325\textwidth}
\includegraphics[width=1.0\linewidth]{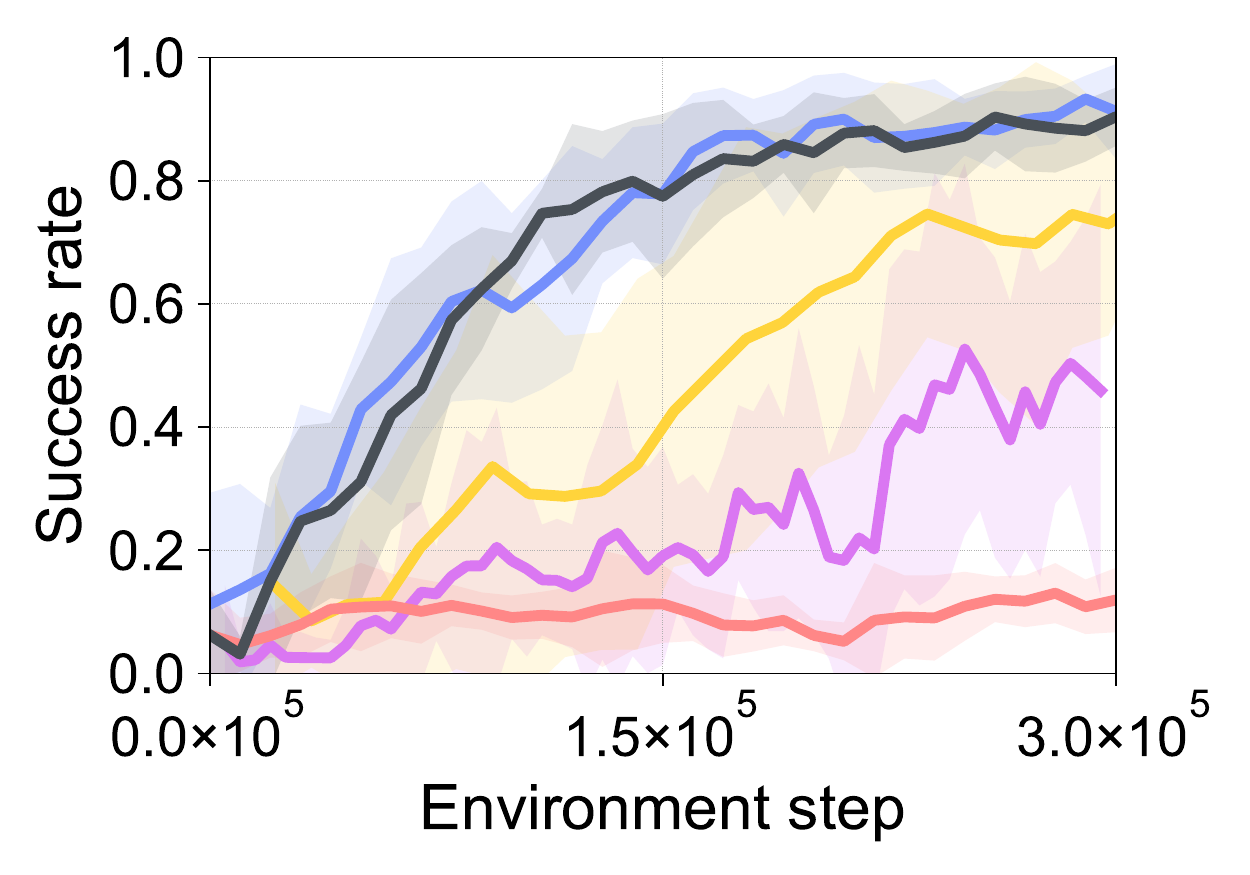}
\caption{Pusher}
\end{subfigure}

\begin{subfigure}{0.325\textwidth}
\includegraphics[width=1.0\linewidth]{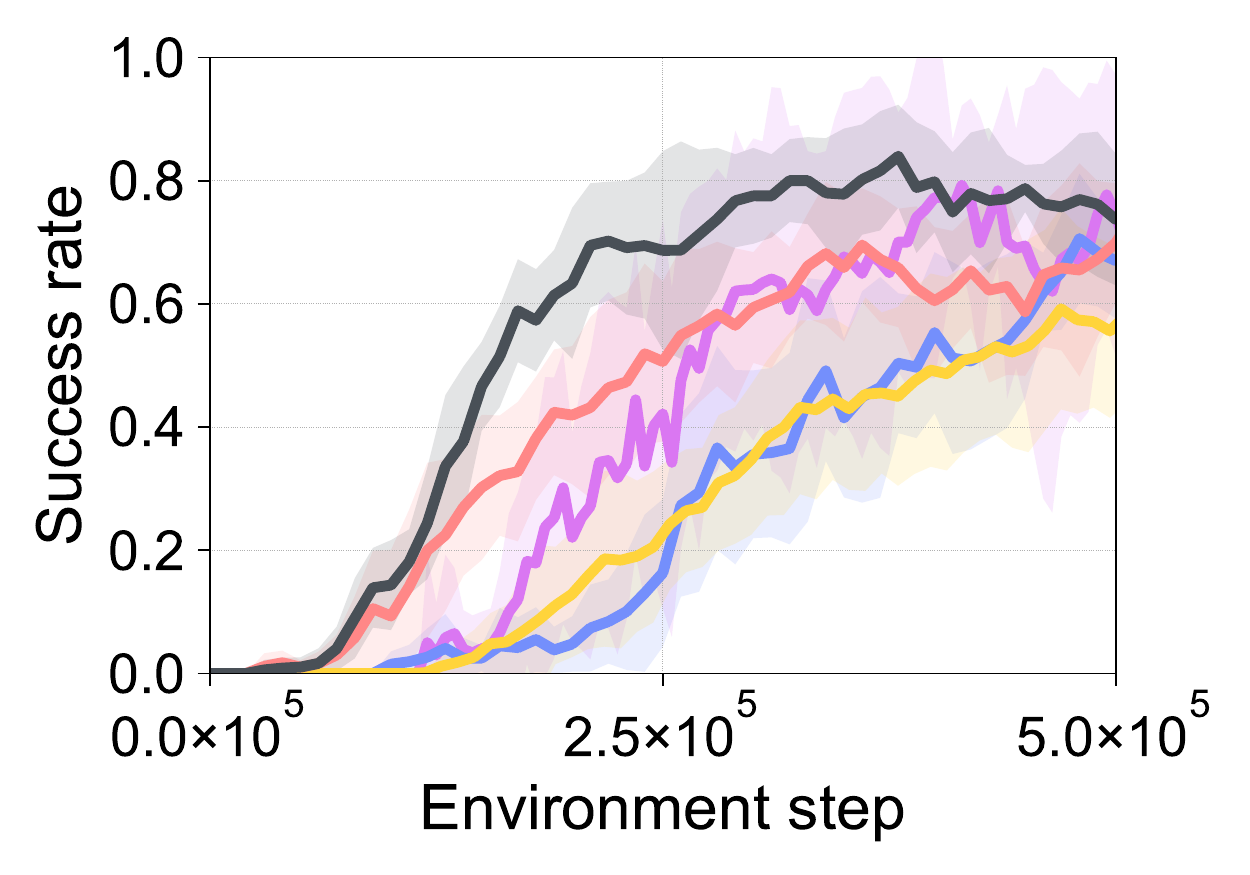}
\caption{L-shaped AntMaze}
\end{subfigure}
\begin{subfigure}{0.325\textwidth}
\includegraphics[width=1.0\linewidth]{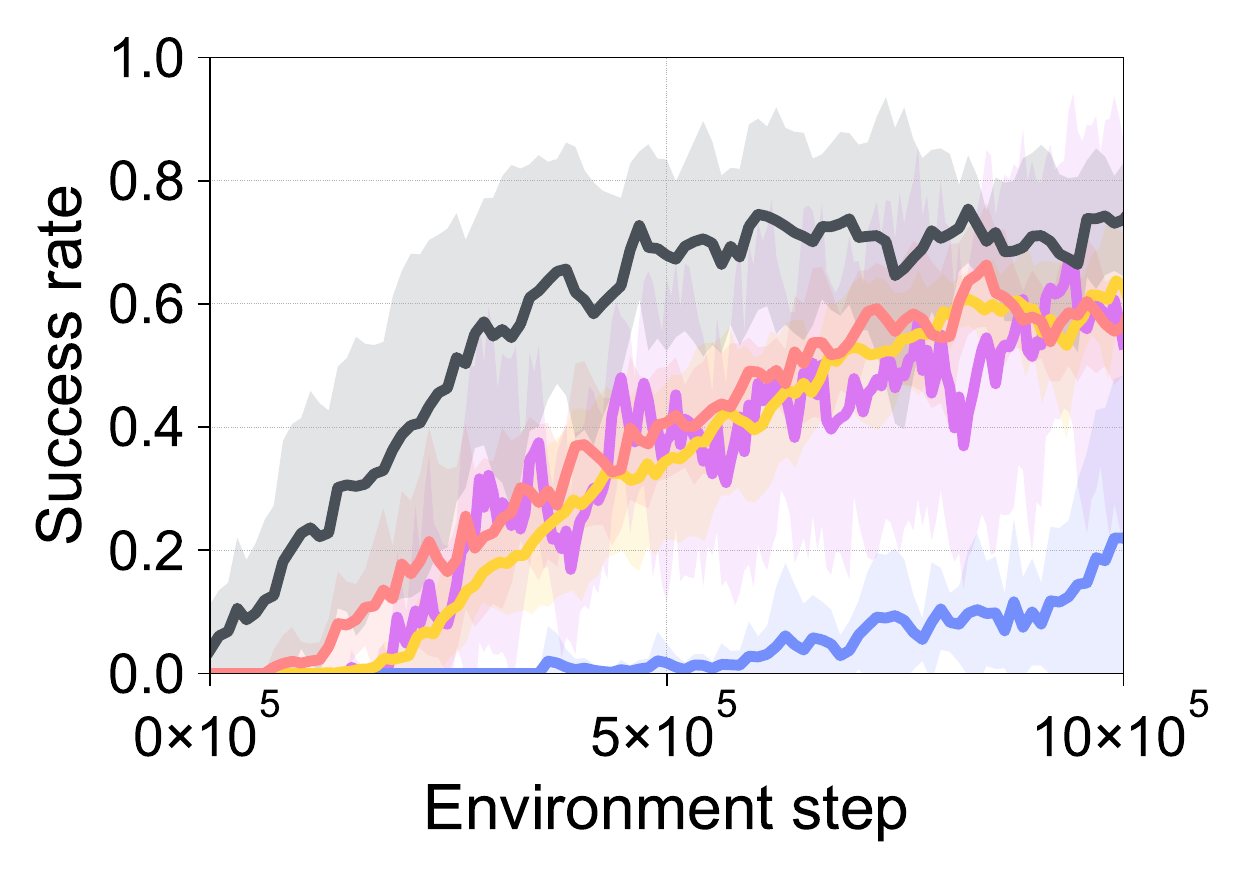}
\caption{U-shaped AntMaze}
\end{subfigure}
\begin{subfigure}{0.325\textwidth}
\includegraphics[width=1.0\linewidth]{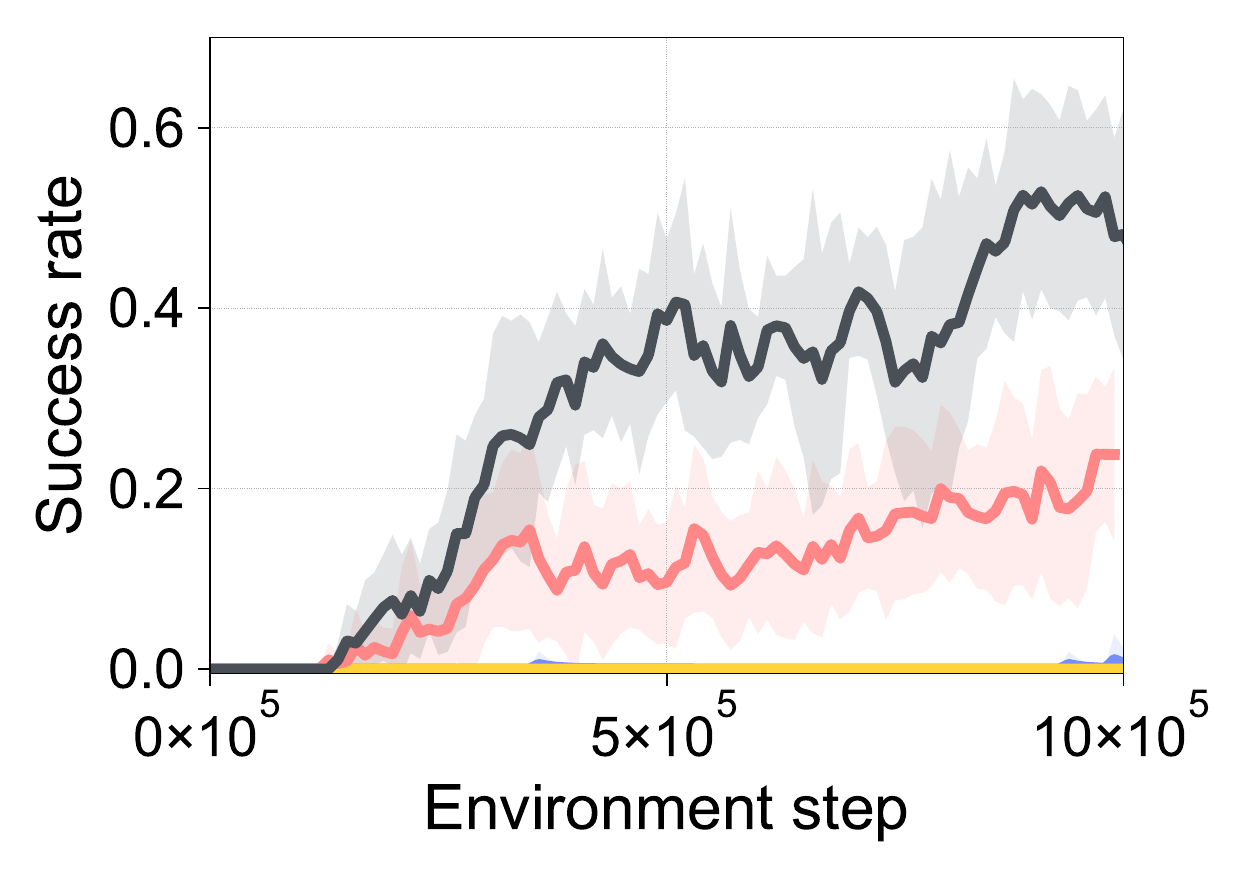}
\caption{Large U-shaped AntMaze}
\end{subfigure}

\begin{subfigure}{0.325\textwidth}
\includegraphics[width=1.0\linewidth]{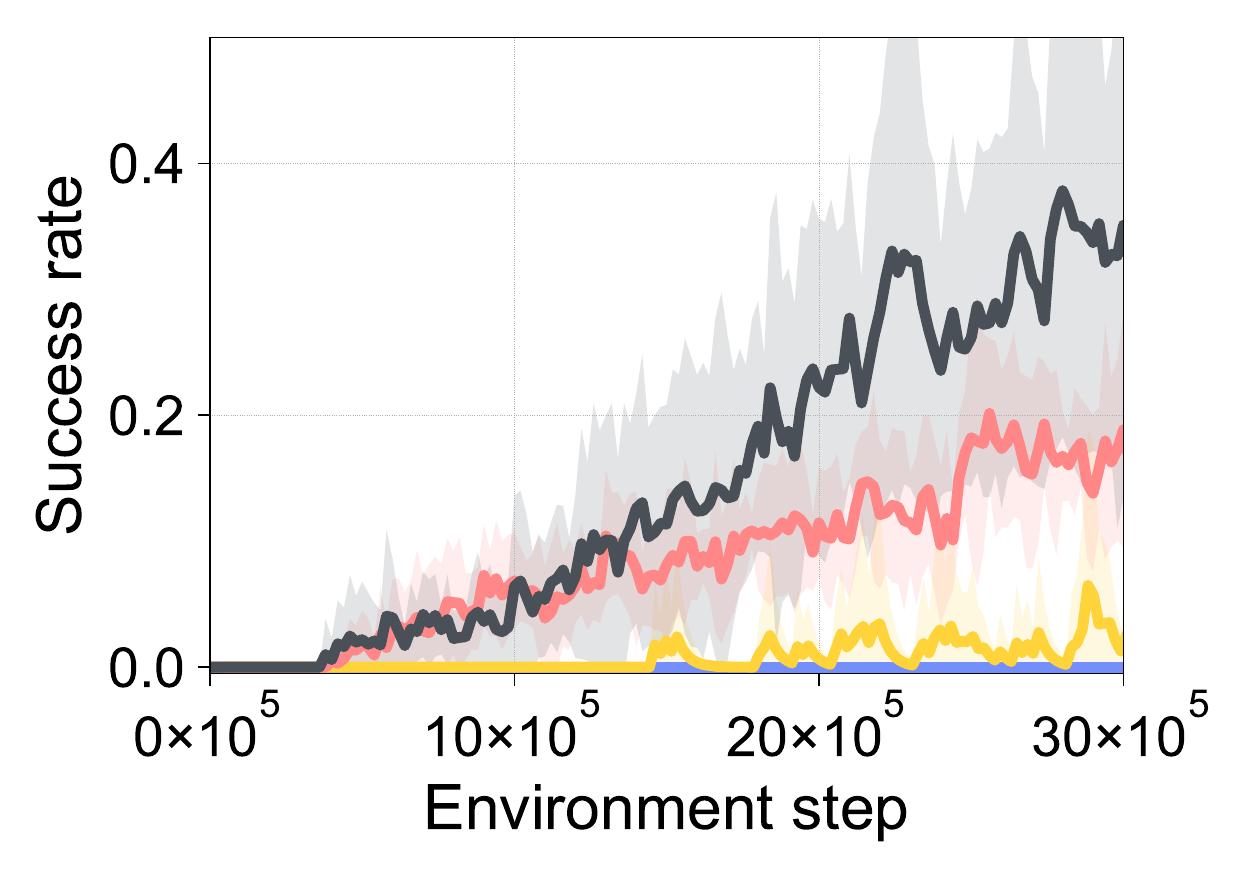}
\caption{S-shaped AntMaze}
\end{subfigure}
\begin{subfigure}{0.325\textwidth}
\includegraphics[width=1.0\linewidth]{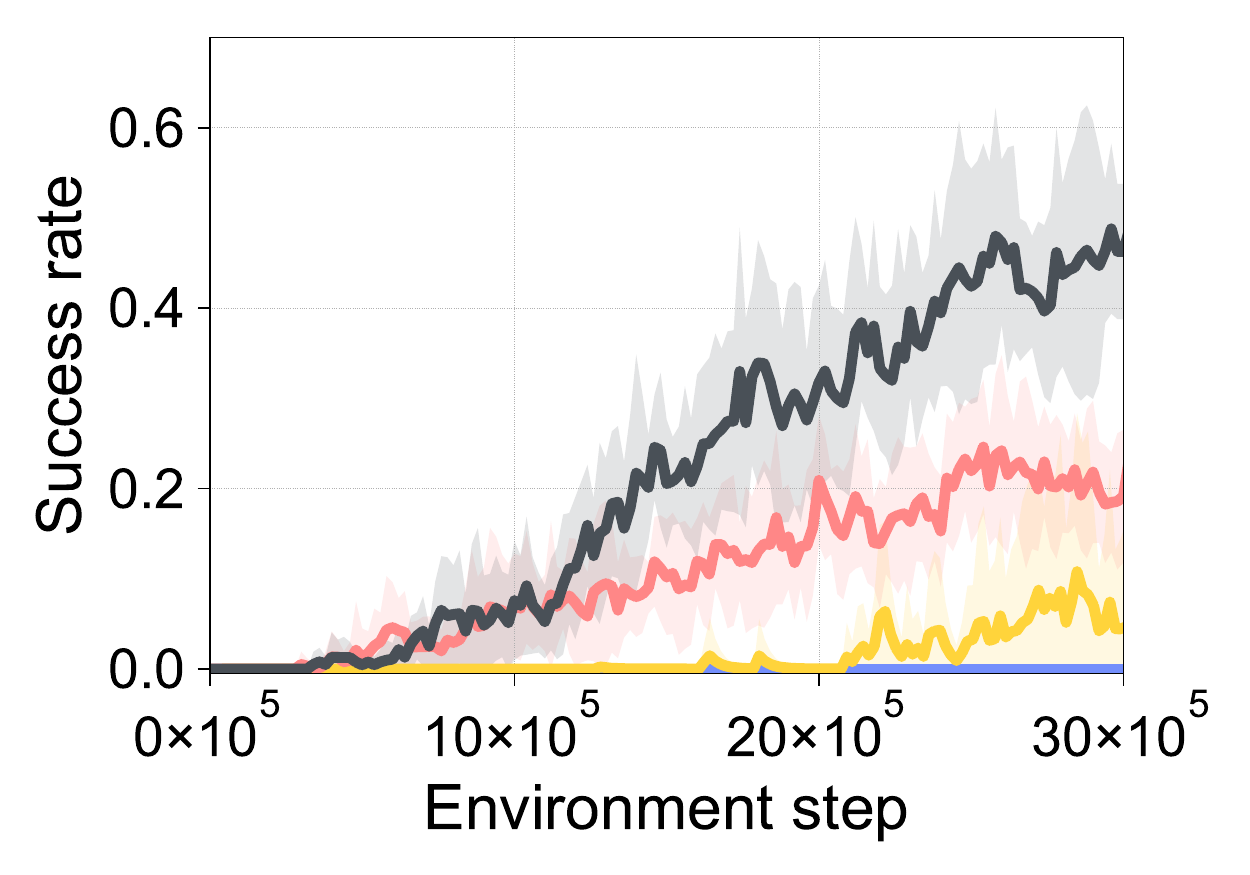}
\caption{$\omega$-shaped AntMaze}
\end{subfigure}
\begin{subfigure}{0.325\textwidth}
\includegraphics[width=1.0\linewidth]{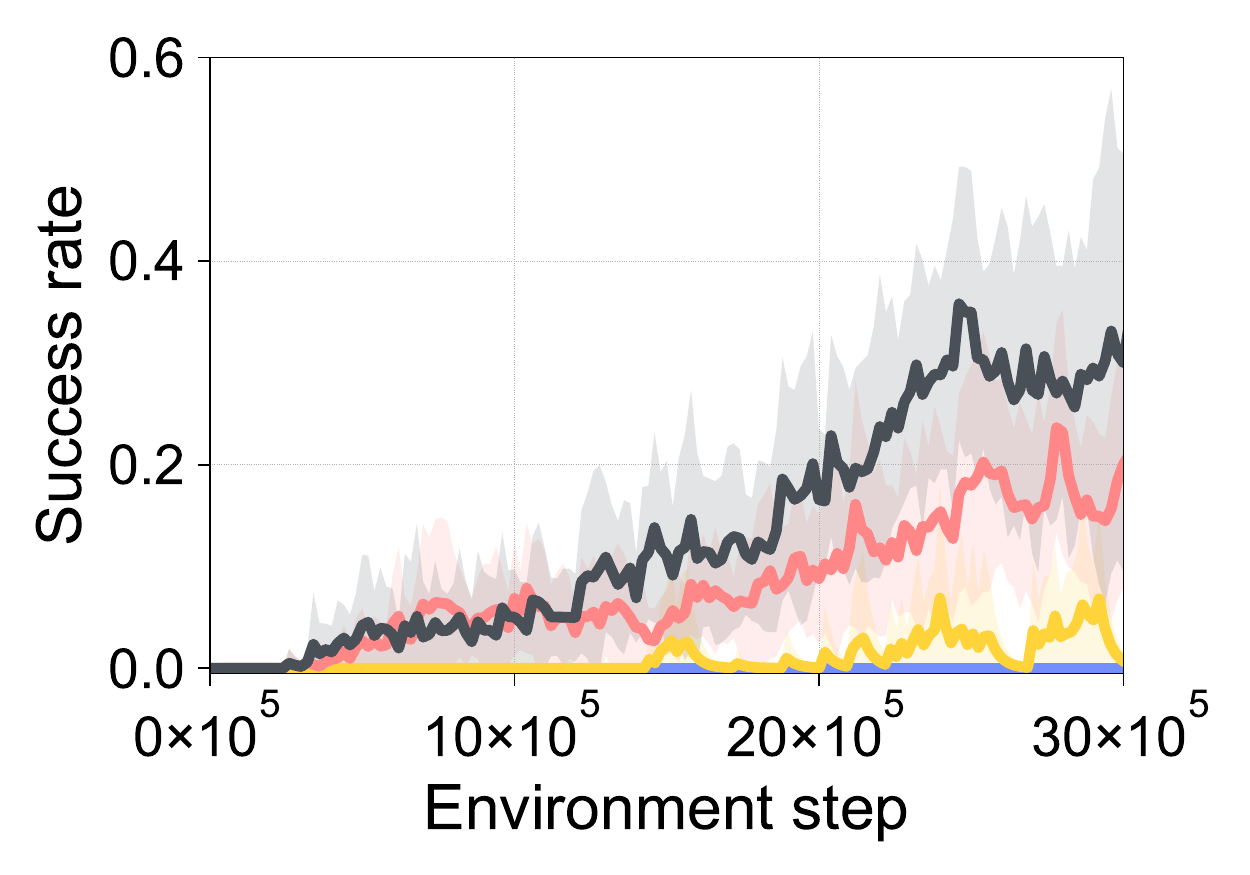}
\caption{$\Pi$-shaped AntMaze}
\end{subfigure}

\caption{
Learning curves on various continuous control tasks as measured on the success rate. We report mean and standard deviation, which are represented as solid lines and shaded regions, respectively, over eight runs.
We observe that \ALGname significantly improves the sample-efficiency of MSS on most tasks.
Note that HER and HIGL perform success rate of 0 for Large U-, S-, $\omega$-, and $\Pi$- shaped AntMaze, and \LthreeP performs success rate of 0 for Large U-shaped AntMaze.
GCSL performs success rate of 0 because it does not use planner in execution, which boosts performance in complex long-horizon tasks, so we compare ours with GCSL-variant that uses a planner 
in Figure~\ref{fig:ablation}. 
}
\label{fig:main}
\end{figure*}
We conduct our experiments on a set of challenging long-horizon continuous control tasks based on MuJoCo simulator \citep{todorov2012mujoco}.
Specifically, we evaluate our framework on 2DReach, Reacher, Pusher, and $\{\text{L}, \text{U}, \text{S},\omega,\Pi\}$-shaped AntMaze environments (see Figure~\ref{fig:environment} for the visualization of environments).
In 2DReach and AntMaze environments, we use a pre-defined 2-dimensional goal space that represents the $(x, y)$ position of the agent following prior works \citep{huang2019mapping, kim2021landmark}. For Reacher, the goal space is 3-dimension that represents the position of an end-effector. For Pusher, the goal space is 6-dimension that represents the positions of an end-effector and a puck. We provide more details of the environments in Supplemental material~\ref{supp:env}.

\textbf{Implementation.} We use DDPG algorithm \citep{Lillicrap2015continuous} as an underlying RL algorithm following the prior work \citep{huang2019mapping}. For a graph-based planner and hindsight goal-relabelling strategy, we follow the setup in MSS \citep{huang2019mapping}. 
We provide more details of the implementation, including the graph-construcion and hyperparameters in Supplemental material~\ref{supp:impl}.

\textbf{Evaluation.}
We run 10 test episodes without an exploration factor for every 50 training episodes. 
For the performance metric, we report the success rate defined as the fraction of episodes where the agents succeed in reaching the target-goal within a threshold.
We report mean and standard deviation, which are represented as solid lines and shaded regions, respectively, over eight runs for Figure~\ref{fig:main} and four runs for the rest of the experiments.
For visual clarity, we smooth all the curves equally.

\begin{figure*}
    \vspace{-.2in}
    \centering
    \begin{subfigure}{0.325\textwidth}
    \includegraphics[width=1.0\linewidth]{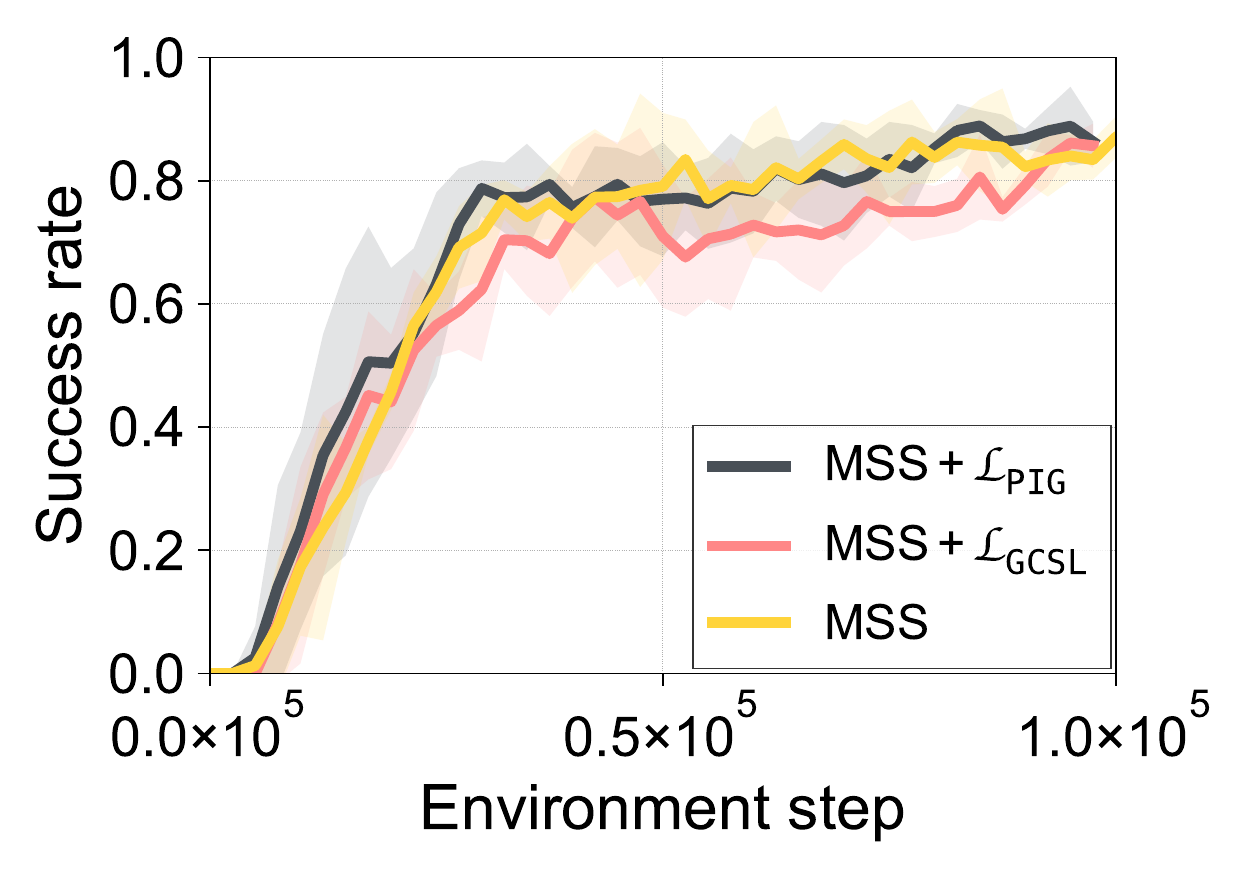}
    \caption{2DReach}
    \label{fig:bc_2dreach}
    \end{subfigure}
    \begin{subfigure}{0.325\textwidth}
    \includegraphics[width=1.0\linewidth]{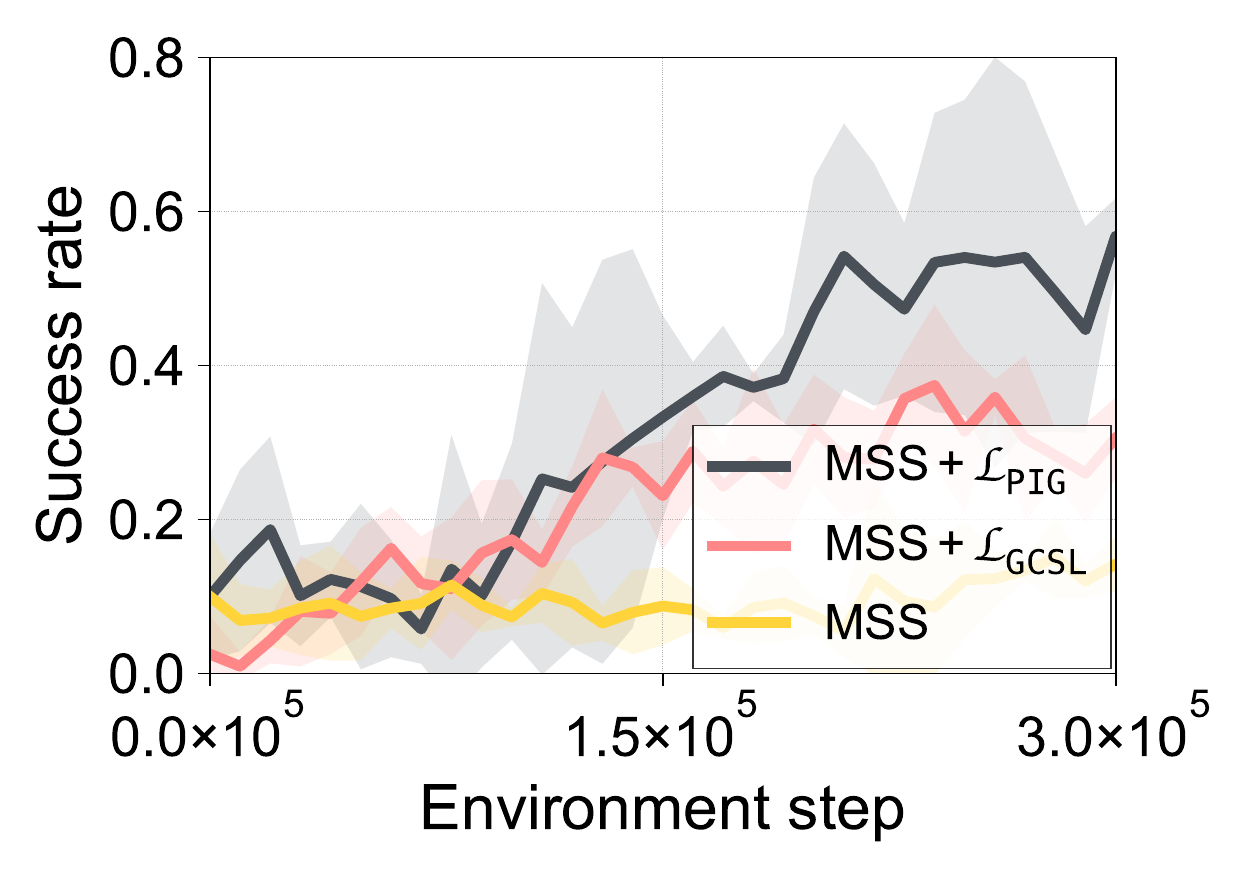}
    \caption{Pusher}
    \label{fig:bc_pusher}
    \end{subfigure}
    \begin{subfigure}{0.325\textwidth}
    \includegraphics[width=1.0\linewidth]{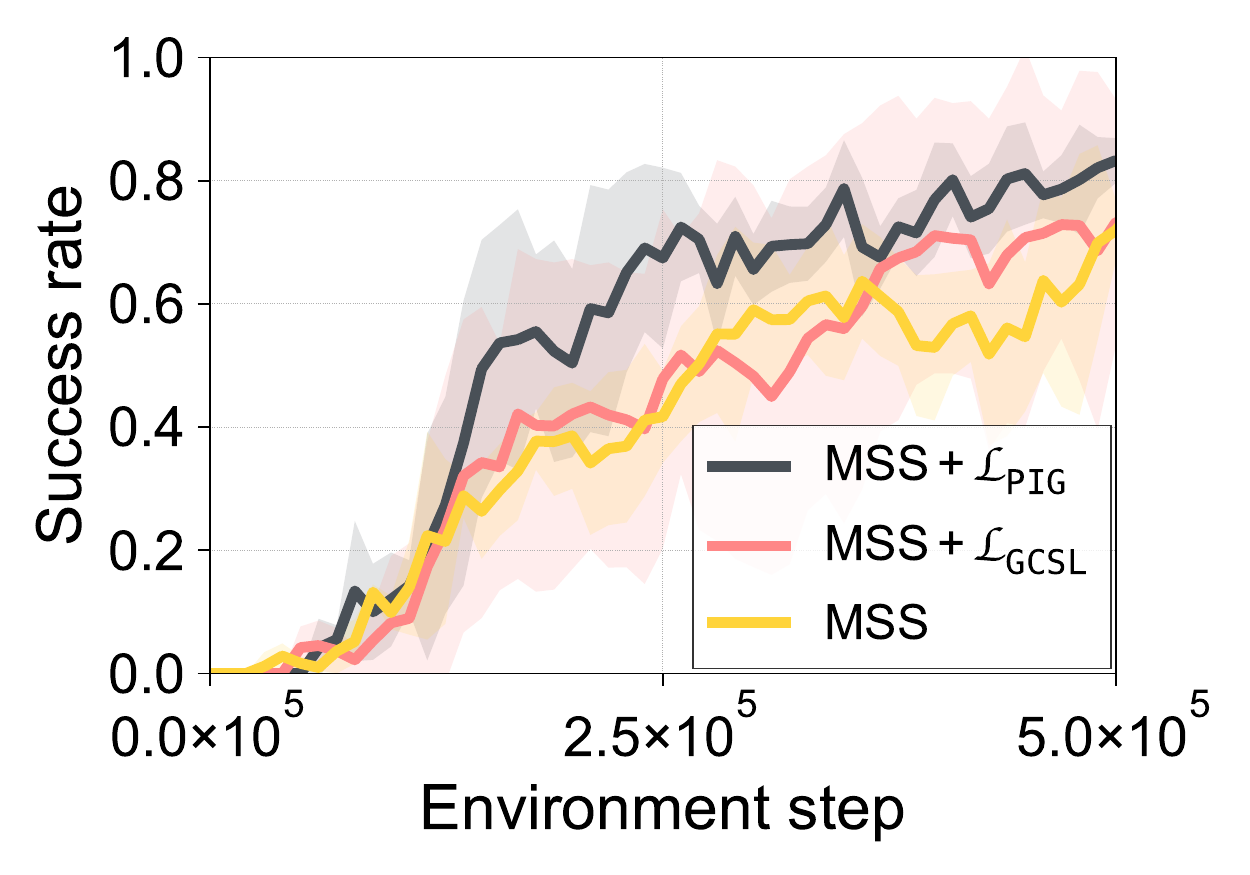}
    \caption{L-shaped AntMaze}
    \label{fig:bc_antmazel}
    \end{subfigure}
    \caption{Ablation studies about self-imitation learning for training on (a) 2DReach, (b) Pusher, and (c) L-shaped Ant Maze with four runs.
    MSS + $\mathcal{L}_{\mathtt{\ALGname}}$ and MSS + $\mathcal{L}_{\mathtt{GCSL}}$ refer to an algorithm that applies loss term $\mathcal{L}_{\mathtt{\ALGname}}$ and $\mathcal{L}_{\mathtt{GCSL}}$ on top of MSS method, respectively; subgoal skipping is not applied. We find that our loss term $\mathcal{L}_{\mathtt{\ALGname}}$ is more effective than $\mathcal{L}_{\mathtt{GCSL}}$ as an auxiliary term. 
    }
    \label{fig:ablation}
\end{figure*}

\textbf{Baselines and our framework.}
We compare our framework with the following baselines on the environments of continuous action spaces:
\begin{itemize}[leftmargin=5.5mm]
    \item HER \citep{andrychowicz2017hindsight}: This method does not use a planner and trains a non-hierarchical policy using a hindsight goal-relabeling strategy.
    \item MSS \citep{huang2019mapping}: This method collects samples using a graph-based planner along with a policy and trains the policy using stored transitions with goal-relabeling by HER. A graph is built via farthest point sampling \citep{vassilvitskii2006k} on states stored in a replay buffer.
    \item 
    \LthreeP \citep{zhang2021world}:
    When building a graph, this method replaces the farthest point sampling of MSS with node-sampling on learned latent space, where nodes are scattered in terms of reachability estimates.
    \item HIGL \citep{kim2021landmark}: This method utilizes a graph-based planner to guide training a high-level policy in goal-conditioned hierarchical reinforcement learning. In contrast, \ALGname uses the planner to guide low-level policy. Comparison with HIGL evaluates the benefits of directly transferring knowledge from the planner to low-level policy without going through high-level policy.
    \item GCSL \citep{ghosh2021learning}: This method learns goal-conditioned policy via iterative supervised learning with goal-relabeling. Originally, GCSL does not utilize a graph-based planner, but we compare ours with GCSL-variant that uses the planner for further investigation in Figure~\ref{fig:ablation}.
\end{itemize}
For all experiments, we report the performance of \ALGname combined with MSS.
Nevertheless, we remark that our work is also compatible with other GCRL approaches because \ALGname does not depend on specific graph-building or planning algorithms, 
as can be seen in Algorithm~\ref{alg:framework} in Supplemental material~\ref{supp:algo_table}.
We provide more details about baselines in Supplemental material~\ref{supp:hyperparameters}.

\subsection{Comparative evaluation}
As shown in Figure~\ref{fig:main}, applying our framework on top of the existing GCRL method, MSS + \ALGname, improves sample-efficiency with a significant margin across various control tasks. Specifically, MSS + \ALGname achieves a success rate of 57.41\% in large U-shaped AntMaze at environment step $10 \times 10^{5}$, while MSS performs 19.08\%. We emphasize that applying \ALGname is more effective when the task is more difficult; MSS + \ALGname shows a larger margin in performance in more difficult tasks (i.e., U-, S-, and $\omega$- shaped mazes rather than L-shaped mazes).
Notably, we also observe that MSS + \ALGname outperforms \LthreeP, which shows that our method can achieve strong performance without the additional complexity of learning latent landmarks.
We remark that \ALGname is also compatible with other GCRL approaches, including \LthreeP, as our framework is agnostic to how graphs are constructed. 
To further support this, we provide additional experimental results that apply \ALGname on top of another graph-based GCRL method in Supplemental material~\ref{supp:l3p_pig}.

\begin{wrapfigure}{r}{0.4\linewidth}
\vspace{0.2in}
     \vspace{-.4in}
    \centering
    \includegraphics[width=1.0\linewidth]{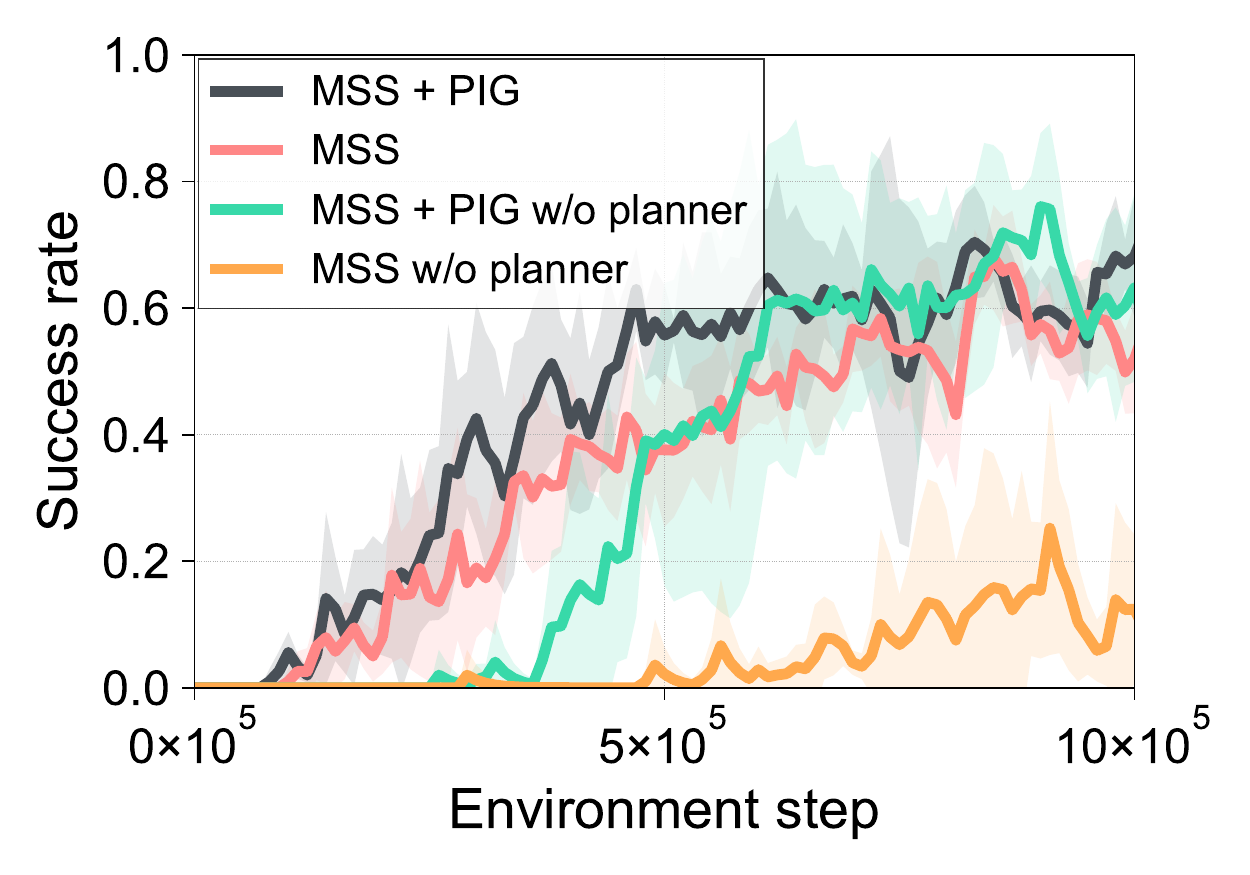}
    \caption{Test time success rate of \ALGname and MSS on U-shaped Ant Maze environment over four runs. The \textbf{w/o planner} means that planner is not used at test time, so a goal is directly fed into the policy instead of a subgoal.
    }
    \label{fig:planner}
\end{wrapfigure}

\label{sec:higl}
Also, we find that MSS + \ALGname outperforms HIGL in Figure~\ref{fig:main}. 
These results show that transferring knowledge from a planner to low-level policy is more efficient than passing through a high-level policy. 
Nevertheless, one can guide both high- and low- level policy via planning, i.e., 
HIGL + \ALGname, which would be interesting future work.
We also remark that the overhead of applying \ALGname is negligible in time complexity. Specifically, both the graph-based planning algorithms (MSS+\ALGname and MSS) spend 1h 30m for 500k steps of the 2DReach, while non-planning baseline (HER) spends 1h.

\textbf{Reaching a goal without a planner at test time.}
To further investigate whether knowledge from graph-based planning is transferred into a policy, we additionally evaluate without the planner at the test time; in other words, the planner is only used at the training time and not anymore at the test time.
Intriguingly, we find that training with our \ALGname enables successfully reaching the target-goal even without the planner at test time. 
As shown in Figure~\ref{fig:planner}, this supports our training scheme indeed makes the policy much stronger.
Such deployment without planning could be practical in some real-world scenarios where a planning time or memory for storing a graph matter \citep{bency2019neural, qureshi2019motion}. We also provide experimental results with a larger maze in Supplemental material~\ref{supp:larger_maze_without_planner}.

\subsection{Ablation studies}
\label{sec:bc_loss}
\textbf{Effectiveness of our loss design.}
In order to empirically demonstrate that utilizing (a) the graph-based planner and (b) actions from a current policy is crucial,
we compare \ALGname (without subgoal skipping) to a GCSL-variant\footnote{Original GCSL use only $\mathcal{L}_{\mathtt{GCSL}}$, not RL loss term and does not use a planner in execution.} that optimizes the following auxiliary objective in conjunction with the RL objective of MSS framework:
\begin{align}
\label{eq:bc}
\mathcal{L_{\mathtt{GCSL}}} = \mathbb{E}_{(s,a,g) \sim \mathcal{B}} [\Vert \pi_{\theta}(s, g) - a\Vert_{2}^{2}],
\end{align}
that is, it encourages a goal-conditioned policy to imitate previously successful actions to reach a (relabeled) goal; a goal and a reward is relabeled in hindsight. In execution time, we also apply a graph-based planner to GCSL-variant for a fair comparison.
As shown in Figure~\ref{fig:ablation}, \ALGname is more effective than using the loss $\mathcal{L}_{\mathtt{GCSL}}$ in terms of sample-efficiency due to (a) knowledge transferred by a planner and (b) more plausible actions from a current policy (rather than an old policy). 

\begin{figure*}[t]
    \vspace{-.2in}
    \centering
    \begin{subfigure}{0.325\textwidth}
    \includegraphics[width=1.0\linewidth]{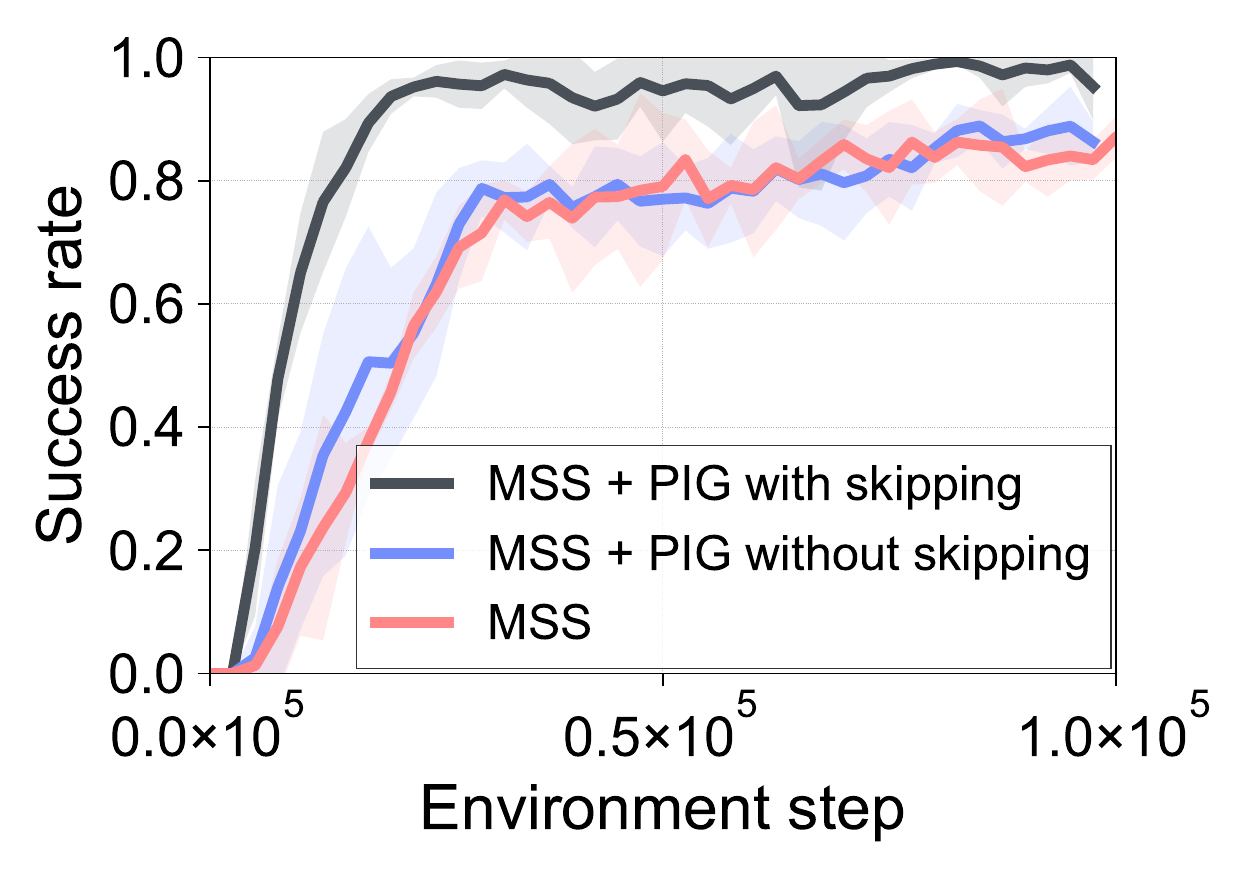}
    \caption{2DReach}
    \end{subfigure}
    \begin{subfigure}{0.325\textwidth}
    \includegraphics[width=1.0\linewidth]{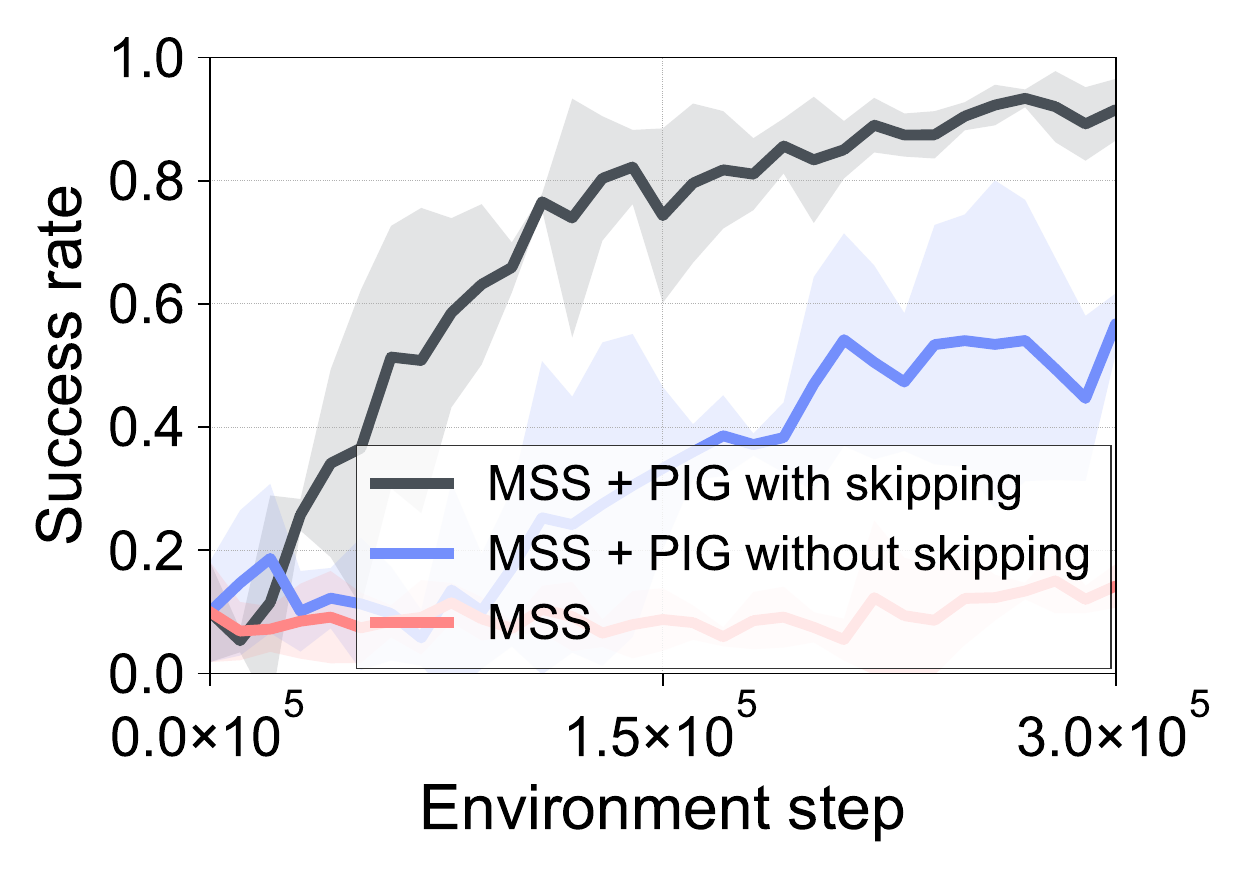}
    \caption{Pusher}
    \end{subfigure}
    \begin{subfigure}{0.325\textwidth}
    \includegraphics[width=1.0\linewidth]{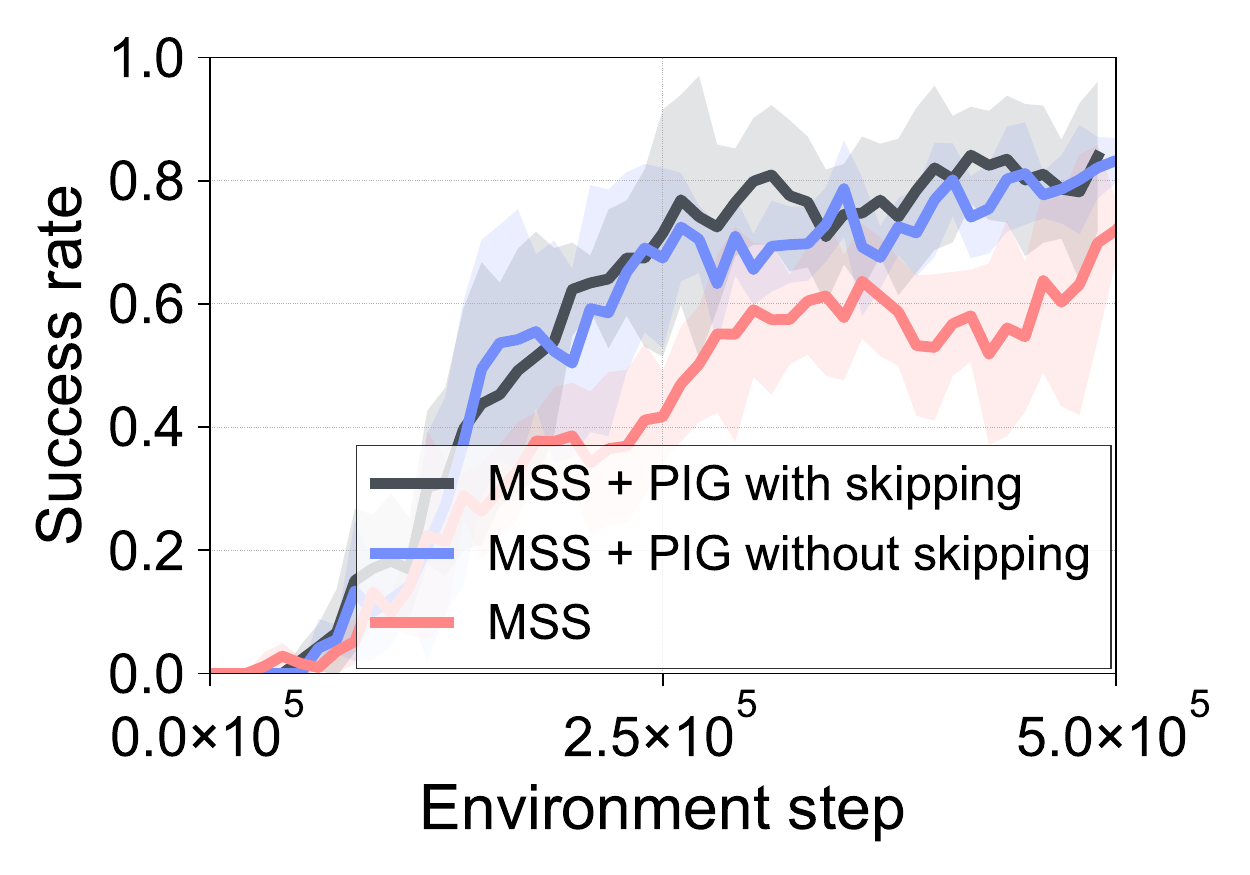}
    \caption{L-shaped AntMaze}
    \end{subfigure}
    \caption{Learning curves of PIG with and without subgoal skipping on (a) 2DReach, (b) Pusher, and (c) L-shaped AntMaze tasks with four runs. \ALGname with subgoal skipping achieves significantly better performance than without skipping in 2DReach and Pusher.}
    \label{fig:skipping}
\end{figure*}
\textbf{Subgoal skipping.}
\label{sec:exp_subgoal_skipping}
We evaluate whether the proposed subgoal skipping is effective in Figure~\ref{fig:skipping}. For 2DReach and Pusher, we observe that \ALGname with skipping achieves significantly better performance than without skipping. 
We understand this is because a strong policy may find a better goal-reaching path by ignoring some of the subgoals proposed by the planner.
On the other hand, we find that subgoal skipping does not provide a large gain on L-shaped Antmaze, which is a more complex environment. We conjecture that this is because learning a strong policy with high-dimensional state inputs of quadruped ant robots is much more difficult. Nevertheless, we believe this issue can be resolved when the base RL algorithm is 
more improved.
We provide more experiments related to subgoal skipping (i.e., comparison to random skipping) in Supplemental material~\ref{supp:random_skipping}, \ref{supp:tuning_cost}, and \ref{supp:subgoal_expl}.

\begin{wrapfigure}{r}{0.325\linewidth}
     \vspace{-.2in}
    \centering
    \includegraphics[width=1.0\linewidth]{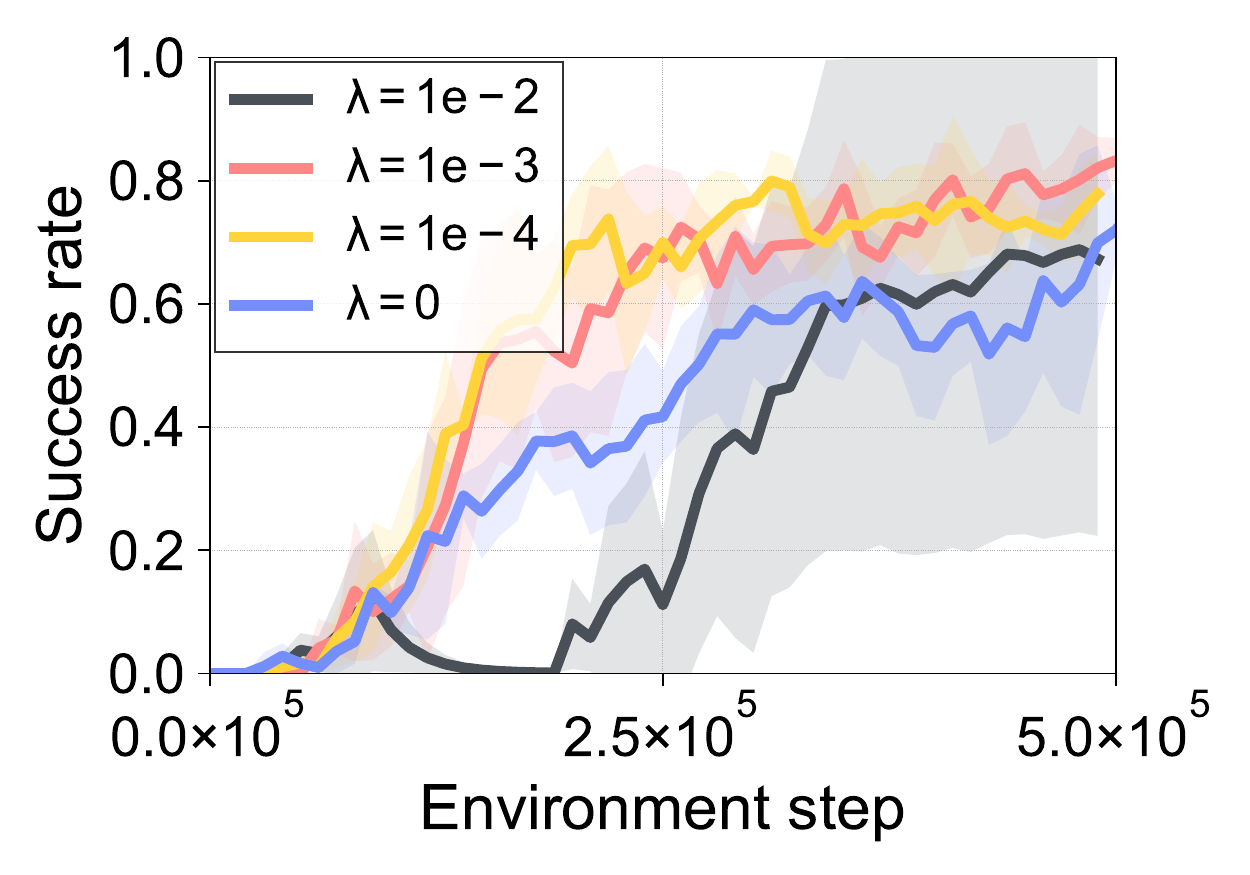}
    \caption{
    Effectiveness of varying balancing coefficient $\lambda$ on L-shaped AntMaze.
    }
    \label{fig:lambda}
    \vspace{-.2in}
\end{wrapfigure}

\textbf{Balancing coefficient $\lambda$.}
We investigate how the balancing coefficient $\lambda$ in Equation~\ref{eq:total_loss} that determines the effect of our proposed loss term $\mathcal{L}_{\mathtt{\ALGname}}$ affect the performance in Figure~\ref{fig:lambda}.
We find that \ALGname with $\lambda \in \{1e-3, 1e-4\}$ outperforms \ALGname with $\lambda = 0$, which shows the importance of the proposed loss.
We also observe that too large value of $\lambda$ harms the performance since it incapacitates the training signal of $\mathcal{L}_{\mathtt{actor}}$ excessively. Meanwhile, one can set the balancing coefficient $\lambda$ automatically in a task-agnostic way, which would guide researchers when they extend our work into new environments in the future.
We provide experimental results with automatic setting of $\lambda$ in Supplemental material~\ref{supp:tuning_cost}.

\vspace{-0.1in}
\section{Conclusion}
We present \ALGname, a new self-improving framework that boosts the sample-efficiency in goal-conditioned RL. We remark that \ALGname is the first work that proposes to guide training and execute with faithfully leveraging the optimal substructure property. Our main idea is (a) distilling planned-subgoal-conditioned policies into the target-goal-conditioned policy and (b) skipping subgoals stochastically in execution based on our loss term. We show that \ALGname on top of the existing GCRL frameworks enhances sample-efficiency with a significant margin across various 
control tasks.
Moreover, based on our findings that a policy could internalize the knowledge of a planner (e.g., reaching a target-goal without a planner), we expect that such a strong policy would enjoy better usage for the scenarios of transfer learning and domain generalization, which we think an interesting future direction.

\textbf{Limitation.} While our experiments demonstrate the \ALGname on top of graph-based goal-conditioned RL method is effective for solving complex control tasks, we only consider the setup where the state space of an agent is a (proprioceptive) compact vector (i.e., state-based RL) following prior works \citep{andrychowicz2017hindsight, huang2019mapping, zhang2021world}. 
In principle, \ALGname is applicable to environments with high-dimensional state spaces because our algorithmic components (self-imitation loss and subgoal skipping) do not depend on the dimensions of state spaces.
It would be interesting future work to extend our work into more high-dimensional observation space such as visual inputs.
We expect that combining subgoal representation learning \citep{nachum2018near, li2021learning} (orthogonal methodology to \ALGname) would be promising.

\section*{Reproducibility statement}
We provide the implementation details of our method in Section~\ref{sec:experiment} and Supplemental material~\ref{supp:impl}. We also open-source our codebase.

\section*{Ethics statement}
This work would promote the research in the field of goal-conditioned RL. However, there goal-conditioned RL algorithms could be misused; for example, malicious users could develop autonomous agents that harm society by setting a dangerous goal.
Therefore, it is important to devise an method that can take consideration of the consequence of its behaviors to a society.

\section*{Acknowledgments and Disclosure of Funding}
We thank Sihyun Yu, Jaeho Lee, Jongjin Park, Jihoon Tack, Jaeyeon Won, Woomin Song, Subin Kim, and anonymous reviewers for providing helpful feedbacks and suggestions in improving our paper. 
This work was supported by Institute of Information \& communications Technology Planning \& Evaluation (IITP) grant funded by the Korea government(MSIT) (No.2019-0-00075, Artificial Intelligence Graduate School Program(KAIST)).
This work was partly supported by Institute of Information \& communications Technology Planning \& Evaluation (IITP) grant funded by the Korea government(MSIT) (No.2022-0-00953,Self-directed AI Agents with Problem-solving Capability). This work was supported by the National Research Foundation of Korea(NRF) grant funded by the Korea government. (MSIT) (2022R1C1C1013366)

\bibliography{iclr2023_conference}

\begin{thebibliography}{31}
\providecommand{\natexlab}[1]{#1}
\providecommand{\url}[1]{\texttt{#1}}
\expandafter\ifx\csname urlstyle\endcsname\relax
  \providecommand{\doi}[1]{doi: #1}\else
  \providecommand{\doi}{doi: \begingroup \urlstyle{rm}\Url}\fi

\bibitem[Andrychowicz et~al.(2017)Andrychowicz, Wolski, Ray, Schneider, Fong,
  Welinder, McGrew, Tobin, Abbeel, and Zaremba]{andrychowicz2017hindsight}
Marcin Andrychowicz, Filip Wolski, Alex Ray, Jonas Schneider, Rachel Fong,
  Peter Welinder, Bob McGrew, Josh Tobin, Pieter Abbeel, and Wojciech Zaremba.
\newblock Hindsight experience replay.
\newblock \emph{arXiv preprint arXiv:1707.01495}, 2017.

\bibitem[Bency et~al.(2019)Bency, Qureshi, and Yip]{bency2019neural}
Mayur~J Bency, Ahmed~H Qureshi, and Michael~C Yip.
\newblock Neural path planning: Fixed time, near-optimal path generation via
  oracle imitation.
\newblock In \emph{IEEE/RSJ International Conference on Intelligent Robots and
  Systems (IROS)}, 2019.

\bibitem[Chane-Sane et~al.(2021)Chane-Sane, Schmid, and Laptev]{chane2021goal}
Elliot Chane-Sane, Cordelia Schmid, and Ivan Laptev.
\newblock Goal-conditioned reinforcement learning with imagined subgoals.
\newblock In \emph{International Conference on Machine Learning}, 2021.

\bibitem[Ding et~al.(2019)Ding, Florensa, Abbeel, and Phielipp]{ding2019goal}
Yiming Ding, Carlos Florensa, Pieter Abbeel, and Mariano Phielipp.
\newblock Goal-conditioned imitation learning.
\newblock In \emph{Advances in Neural Information Processing Systems}, 2019.

\bibitem[Eysenbach et~al.(2019)Eysenbach, Salakhutdinov, and
  Levine]{eysenbach2019search}
Ben Eysenbach, Russ~R Salakhutdinov, and Sergey Levine.
\newblock Search on the replay buffer: Bridging planning and reinforcement
  learning.
\newblock In \emph{Advances in Neural Information Processing Systems}, 2019.

\bibitem[Ghosh et~al.(2021)Ghosh, Gupta, Reddy, Fu, Devin, Eysenbach, and
  Levine]{ghosh2021learning}
Dibya Ghosh, Abhishek Gupta, Ashwin Reddy, Justin Fu, Coline~Manon Devin,
  Benjamin Eysenbach, and Sergey Levine.
\newblock Learning to reach goals via iterated supervised learning.
\newblock In \emph{International Conference on Learning Representations}, 2021.

\bibitem[Hamrick et~al.(2020)Hamrick, Bapst, Sanchez-Gonzalez, Pfaff, Weber,
  Buesing, and Battaglia]{Hamrick2020Combining}
Jessica~B. Hamrick, Victor Bapst, Alvaro Sanchez-Gonzalez, Tobias Pfaff,
  Theophane Weber, Lars Buesing, and Peter~W. Battaglia.
\newblock Combining q-learning and search with amortized value estimates.
\newblock In \emph{International Conference on Learning Representations}, 2020.

\bibitem[Hoang et~al.(2021)Hoang, Sohn, Choi, Carvalho, and
  Lee]{hoang2021successor}
Christopher Hoang, Sungryull Sohn, Jongwook Choi, Wilka Carvalho, and Honglak
  Lee.
\newblock Successor feature landmarks for long-horizon goal-conditioned
  reinforcement learning.
\newblock In \emph{Advances in Neural Information Processing Systems}, 2021.

\bibitem[Huang et~al.(2019)Huang, Liu, and Su]{huang2019mapping}
Zhiao Huang, Fangchen Liu, and Hao Su.
\newblock Mapping state space using landmarks for universal goal reaching.
\newblock In \emph{Advances in Neural Information Processing Systems}, 2019.

\bibitem[Kaelbling(1993)]{kaelbling1993learning}
Leslie~Pack Kaelbling.
\newblock Learning to achieve goals.
\newblock In \emph{IJCAI}, 1993.

\bibitem[Kim et~al.(2021)Kim, Seo, and Shin]{kim2021landmark}
Junsu Kim, Younggyo Seo, and Jinwoo Shin.
\newblock Landmark-guided subgoal generation in hierarchical reinforcement
  learning.
\newblock In \emph{Advances in Neural Information Processing Systems}, 2021.

\bibitem[Kingma \& Ba(2014)Kingma and Ba]{kingma2014adam}
Diederik~P Kingma and Jimmy Ba.
\newblock Adam: A method for stochastic optimization.
\newblock \emph{arXiv preprint arXiv:1412.6980}, 2014.

\bibitem[Laskin et~al.(2020)Laskin, Emmons, Jain, Kurutach, Abbeel, and
  Pathak]{laskin2020sparse}
Michael Laskin, Scott Emmons, Ajay Jain, Thanard Kurutach, Pieter Abbeel, and
  Deepak Pathak.
\newblock Sparse graphical memory for robust planning.
\newblock \emph{arXiv preprint arXiv:2003.06417}, 2020.

\bibitem[Li et~al.(2021)Li, Zheng, Wang, and Zhang]{li2021learning}
Siyuan Li, Lulu Zheng, Jianhao Wang, and Chongjie Zhang.
\newblock Learning subgoal representations with slow dynamics.
\newblock In \emph{International Conference on Learning Representations}, 2021.

\bibitem[Lillicrap et~al.(2016)Lillicrap, Hunt, Pritzel, Heess, Erez, Tassa,
  Silver, and Wierstra]{Lillicrap2015continuous}
Timothy~P. Lillicrap, Jonathan~J. Hunt, Alexander Pritzel, Nicolas Heess, Tom
  Erez, Yuval Tassa, David Silver, and Daan Wierstra.
\newblock Continuous control with deep reinforcement learning.
\newblock In \emph{International Conference on Learning Representations}, 2016.

\bibitem[Mendonca et~al.(2021)Mendonca, Rybkin, Daniilidis, Hafner, and
  Pathak]{mendonca2021discovering}
Russell Mendonca, Oleh Rybkin, Kostas Daniilidis, Danijar Hafner, and Deepak
  Pathak.
\newblock Discovering and achieving goals via world models.
\newblock In \emph{Advances in Neural Information Processing Systems}, 2021.

\bibitem[Nachum et~al.(2018)Nachum, Gu, Lee, and Levine]{nachum2018data}
Ofir Nachum, Shixiang Gu, Honglak Lee, and Sergey Levine.
\newblock Data-efficient hierarchical reinforcement learning.
\newblock In \emph{Advances in Neural Information Processing Systems}, 2018.

\bibitem[Nachum et~al.(2019)Nachum, Gu, Lee, and Levine]{nachum2018near}
Ofir Nachum, Shixiang Gu, Honglak Lee, and Sergey Levine.
\newblock Near-optimal representation learning for hierarchical reinforcement
  learning.
\newblock In \emph{International Conference on Learning Representations}, 2019.

\bibitem[Oh et~al.(2018)Oh, Guo, Singh, and Lee]{oh2018self}
Junhyuk Oh, Yijie Guo, Satinder Singh, and Honglak Lee.
\newblock Self-imitation learning.
\newblock In \emph{International Conference on Machine Learning}, 2018.

\bibitem[Plappert et~al.(2018)Plappert, Andrychowicz, Ray, McGrew, Baker,
  Powell, Schneider, Tobin, Chociej, Welinder, et~al.]{plappert2018multi}
Matthias Plappert, Marcin Andrychowicz, Alex Ray, Bob McGrew, Bowen Baker,
  Glenn Powell, Jonas Schneider, Josh Tobin, Maciek Chociej, Peter Welinder,
  et~al.
\newblock Multi-goal reinforcement learning: Challenging robotics environments
  and request for research.
\newblock \emph{arXiv preprint arXiv:1802.09464}, 2018.

\bibitem[Pong et~al.(2020)Pong, Dalal, Lin, Nair, Bahl, and
  Levine]{pong2020skew}
Vitchyr Pong, Murtaza Dalal, Steven Lin, Ashvin Nair, Shikhar Bahl, and Sergey
  Levine.
\newblock Skew-fit: State-covering self-supervised reinforcement learning.
\newblock In \emph{International Conference on Machine Learning}, 2020.

\bibitem[Qureshi et~al.(2019)Qureshi, Simeonov, Bency, and
  Yip]{qureshi2019motion}
Ahmed~H Qureshi, Anthony Simeonov, Mayur~J Bency, and Michael~C Yip.
\newblock Motion planning networks.
\newblock In \emph{IEEE International Conference on Robotics and Automation
  (ICRA)}, 2019.

\bibitem[Savinov et~al.(2018)Savinov, Dosovitskiy, and
  Koltun]{savinov2018semiparametric}
Nikolay Savinov, Alexey Dosovitskiy, and Vladlen Koltun.
\newblock Semi-parametric topological memory for navigation.
\newblock In \emph{International Conference on Learning Representations}, 2018.

\bibitem[Schaul et~al.(2015)Schaul, Horgan, Gregor, and
  Silver]{schaul2015universal}
Tom Schaul, Daniel Horgan, Karol Gregor, and David Silver.
\newblock Universal value function approximators.
\newblock In \emph{International Conference on Machine Learning}, 2015.

\bibitem[Silver et~al.(2017)Silver, Schrittwieser, Simonyan, Antonoglou, Huang,
  Guez, Hubert, Baker, Lai, Bolton, et~al.]{silver2017mastering}
David Silver, Julian Schrittwieser, Karen Simonyan, Ioannis Antonoglou, Aja
  Huang, Arthur Guez, Thomas Hubert, Lucas Baker, Matthew Lai, Adrian Bolton,
  et~al.
\newblock Mastering the game of go without human knowledge.
\newblock \emph{nature}, 2017.

\bibitem[Singh et~al.(2003)Singh, Misra, Hnizdo, Fedorowicz, and
  Demchuk]{singh2003nearest}
Harshinder Singh, Neeraj Misra, Vladimir Hnizdo, Adam Fedorowicz, and Eugene
  Demchuk.
\newblock Nearest neighbor estimates of entropy.
\newblock \emph{American Journal of Mathematical and Management Sciences},
  2003.

\bibitem[Sutton \& Barto(2018)Sutton and Barto]{sutton2018reinforcement}
Richard~S Sutton and Andrew~G Barto.
\newblock \emph{Reinforcement learning: An introduction}.
\newblock MIT Press, 2018.

\bibitem[Todorov et~al.(2012)Todorov, Erez, and Tassa]{todorov2012mujoco}
Emanuel Todorov, Tom Erez, and Yuval Tassa.
\newblock Mujoco: A physics engine for model-based control.
\newblock In \emph{IEEE/RSJ International Conference on Intelligent Robots and
  Systems}, 2012.

\bibitem[Vassilvitskii \& Arthur(2006)Vassilvitskii and
  Arthur]{vassilvitskii2006k}
Sergei Vassilvitskii and David Arthur.
\newblock k-means++: The advantages of careful seeding.
\newblock In \emph{ACM-SIAM Symposium on Discrete Algorithms}, 2006.

\bibitem[Zhang et~al.(2021)Zhang, Yang, and Stadie]{zhang2021world}
Lunjun Zhang, Ge~Yang, and Bradly~C Stadie.
\newblock World model as a graph: Learning latent landmarks for planning.
\newblock In \emph{International Conference on Machine Learning}, 2021.

\bibitem[Zhang et~al.(2020)Zhang, Guo, Tan, Hu, and Chen]{zhang2020generating}
Tianren Zhang, Shangqi Guo, Tian Tan, Xiaolin Hu, and Feng Chen.
\newblock Generating adjacency-constrained subgoals in hierarchical
  reinforcement learning.
\newblock In \emph{Advances in Neural Information Processing Systems}, 2020.

\end{thebibliography}
\bibliographystyle{iclr2023_conference}

\appendix
\newpage
\appendix
\section{Algorithm table}
\label{supp:algo_table}
We provide algorithm tables that represent \ALGname in Algorithm \ref{alg:framework} and \ref{alg:skip}.
\begin{algorithm}[h]
   \caption{GCRL with planning + \highlight{\ALGname}}
   \label{alg:framework}
\begin{algorithmic}
\State {\bfseries Input}: Number of training episodes $M$, horizon $H$
\State Initialize replay buffer $\mathcal{B} \leftarrow \varnothing$.
\State Initialize the parameters of goal-conditioned policy $\pi_{\theta}$.
\State Initialize the parameters of action-value function $Q_{\phi}$.
\For{$m=1, 2, 3, \ldots M$}
\State Reset the environment.
\State Sample a target goal $g$ and an initial state $s_{0}$.
    \For{$t=1, 2, 3, \ldots H$}
    \State Build a graph $\mathcal{H} = (\mathcal{V}, \mathcal{E}, d)$ using $\mathcal{B}$.
    \State Find the shortest subgoal-path $\tau_{g}$ from $s_{t}$ to $g$.
    \State \highlight{Find a desired subgoal $l^{*}$ via Algorithm~\ref{alg:skip}.}
    \State Collect a transition $(s_{t}, a_{t}, r_{t})$ using $\pi_{\theta} (s_{t}, l^{*})$.
    \State Store the transition and the planned path $\tau_{g}$ in $\mathcal{B}$.
    \EndFor
\State Update $Q_{\phi}$ using $\mathcal{L}_{\mathtt{critic}} (\phi)$ of Equation~\ref{eq:ddpg_critic}
\State Update $\pi_{\theta}$ using $\mathcal{L}_{\mathtt{actor}} (\theta) + \highlight{\lambda \mathcal{L}_{\mathtt{\ALGname}} (\theta)} $ of Equation~\ref{eq:total_loss}
\EndFor
\end{algorithmic}
\end{algorithm}
\begin{algorithm}[h]
   \caption{Subgoal skipping for execution}
   \label{alg:skip}
\begin{algorithmic}
\State {\bfseries Input}: Subgoal-path $\tau_{g} = (l^{1}, l^{2}, \ldots, l^{N})$, \\ \; \; \; \; \; \;the latest $\mathcal{L}_{\mathtt{\ALGname, latest}}$, normalizing constant $C$.
\State Initialize desired subgoal $l^{*} \leftarrow l^{2}$, current index $i \leftarrow 2$
\While {$i < N$}
    \State Sample $\mathtt{jump}$ according to Equation~\ref{eq:path_hop}. 
    \If{$\mathtt{jump}$}
        \State Update current index $i \leftarrow i+1$.
        \State Update desired subgoal $l^{*} \leftarrow l^{i}$.
    \Else
        \State Break while loop.
    \EndIf
\EndWhile
\State {\bfseries Output}: desired subgoal $l^{*}$
\end{algorithmic}
\end{algorithm}
\newpage

\section{Environment details}
\label{supp:env}
\subsection{2DReach} 
A green point in a 2D U-shaped Maze of size $15 \times 15$ aims to reach a target goal represented by a red point. At each step, the agent can move within $[-1, 1] \times [-1, 1]$ in $x$ and $y$ directions.

\subsection{Reacher}
A robotic arm aims to make its end-effector reach the target position on 3D space. The state of the arm is 17-dimension, including the positions, angles, and velocities of itself, and the action-space is 7-dimension. Initial point and target goal are set randomly at the start of episode both at training and test time.
    
\subsection{Pusher} A robotic arm aims to make a puck in a plane reach a goal position by pushing the object. The state of the arm is 20-dimension, which is same to Reacher but additionally include position of a puck, and the action-space is 7-dimension. Initial point and target goal are set randomly at the start of episode both at training and test time.
    
\subsection{AntMaze} A quadruped ant robot is trained to reach a random goal from a random location and tested under the most difficult setting for each maze. The states of ant is 30-dimension, including positions and velocities. An ant should reach the target point within 500 steps for U-shaped mazes, and 1000 steps for S-, $\omega$-, and $\Pi$-shaped mazes.

\newpage
\section{Additional experiments}
\label{supp:additioanl_experiments}
\subsection{Applying \ALGname on top of another graph-based GCRL method.}
\label{supp:l3p_pig}
\begin{figure}[h]
\centering
    \begin{subfigure}[b]{0.325\columnwidth}
    \includegraphics[width=\linewidth]{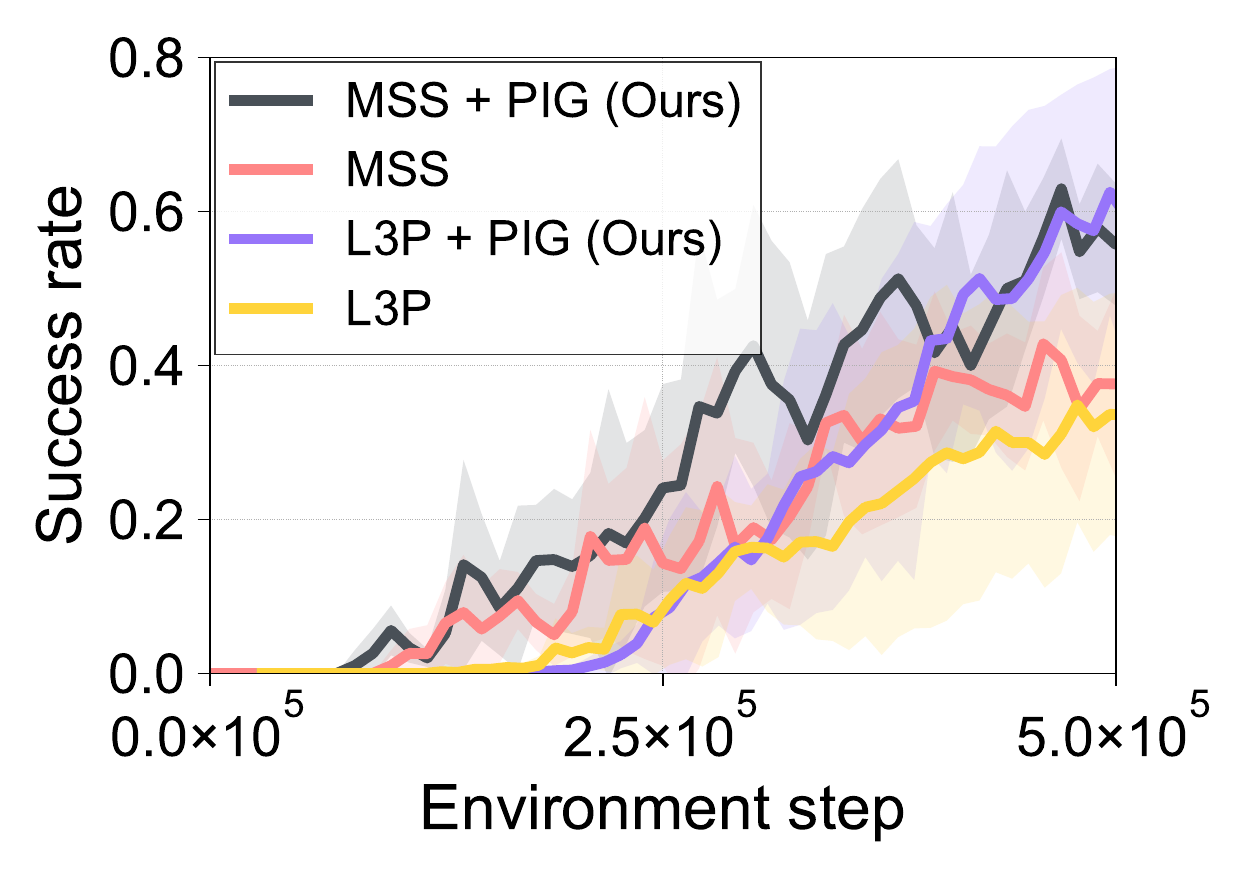}
    \caption{U-shaped AntMaze}
    \label{fig:l3p_antmaze_u}
    \end{subfigure}
\centering
    \begin{subfigure}[b]{0.325\columnwidth}
    \includegraphics[width=\linewidth]{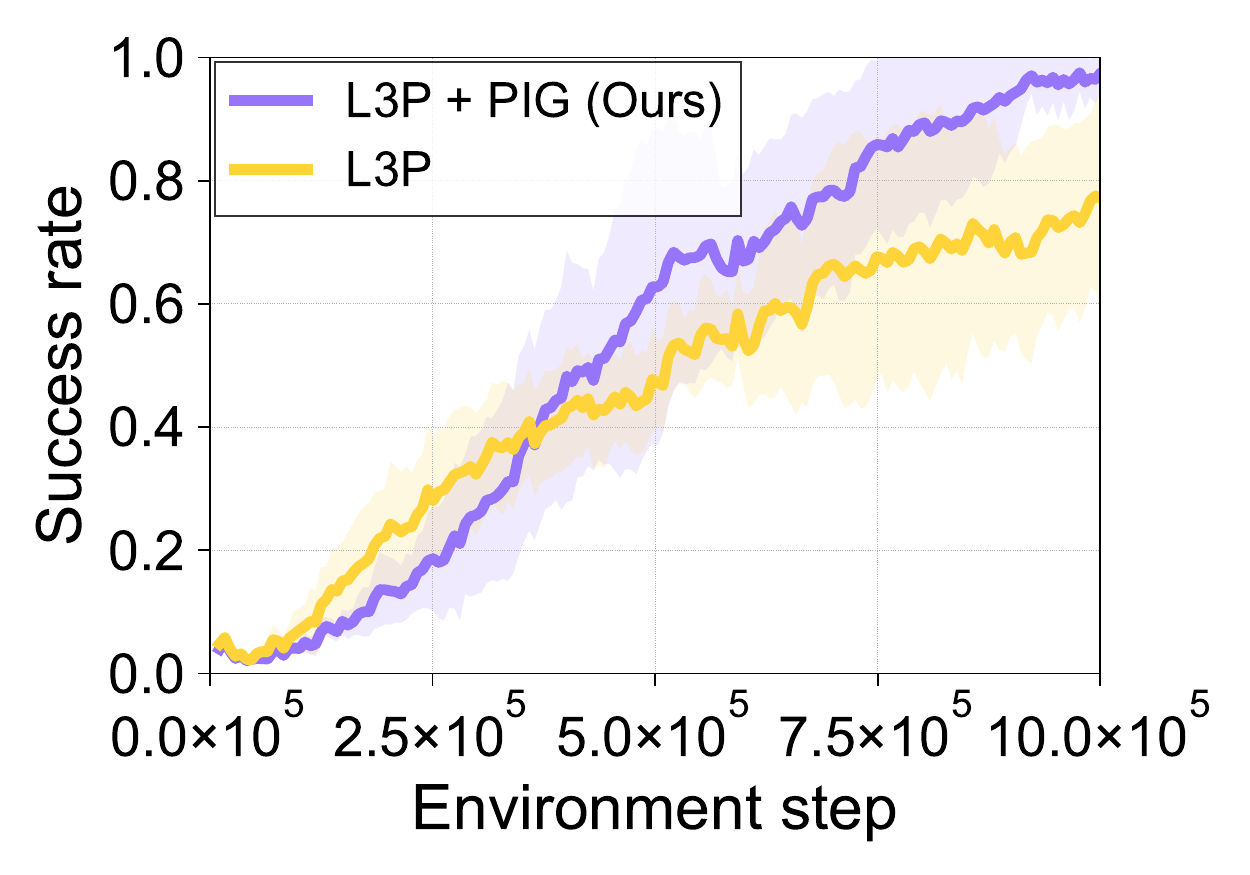}
    \caption{FetchPickAndPlace-v1}
    \label{fig:l3p_pick}
    \end{subfigure}
    \caption{Test time success rate of \ALGname on top of another planning-based GCRL method (i.e., \LthreeP) in (a) U-shaped AntMaze and (b) FetchPickAndPlace-v1.}
    \label{fig:l3p}
\end{figure}

Additionally, we also observe that applying \ALGname on top of another planning-based GCRL method (i.e., \LthreeP rather than MSS) also demonstrates significant gains. As shown in Figure~\ref{fig:l3p}, \ALGname boost sample-efficiency for \LthreeP in U-shaped AntMaze and FetchPickAndPlace-v1~\citep{plappert2018multi}. These experiments further highlight that \ALGname is generic technique to improve performance of all the graph-based planning algorithms. 

\subsection{Comparison to alternatives for subgoal skipping.}
\label{supp:random_skipping}
\begin{figure}[h]
\centering
    \begin{subfigure}[b]{0.325\columnwidth}
    \includegraphics[width=\linewidth]{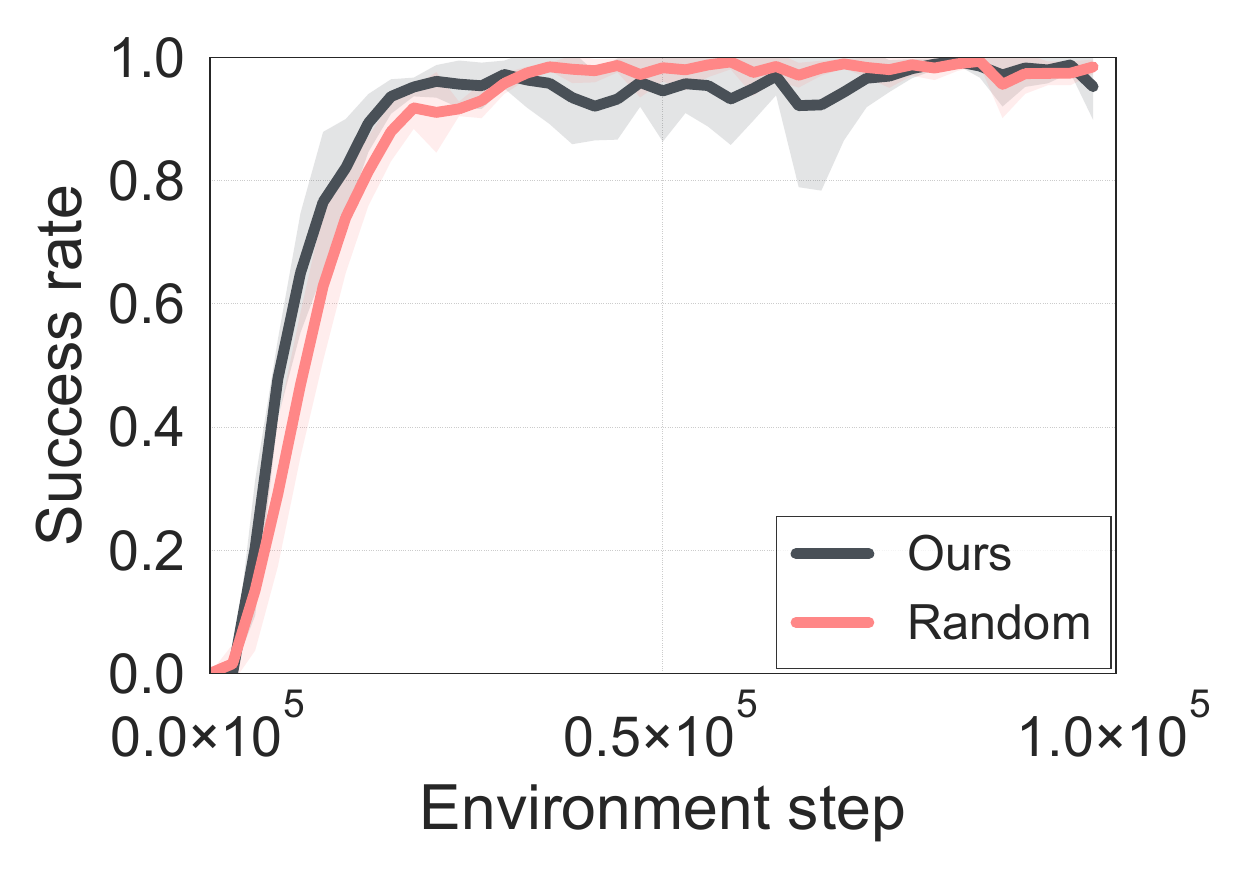}
    \caption{2DReach}
    \label{fig:2dreach_random_skip}
    \end{subfigure}
\centering
    \begin{subfigure}[b]{0.325\columnwidth}
    \includegraphics[width=\linewidth]{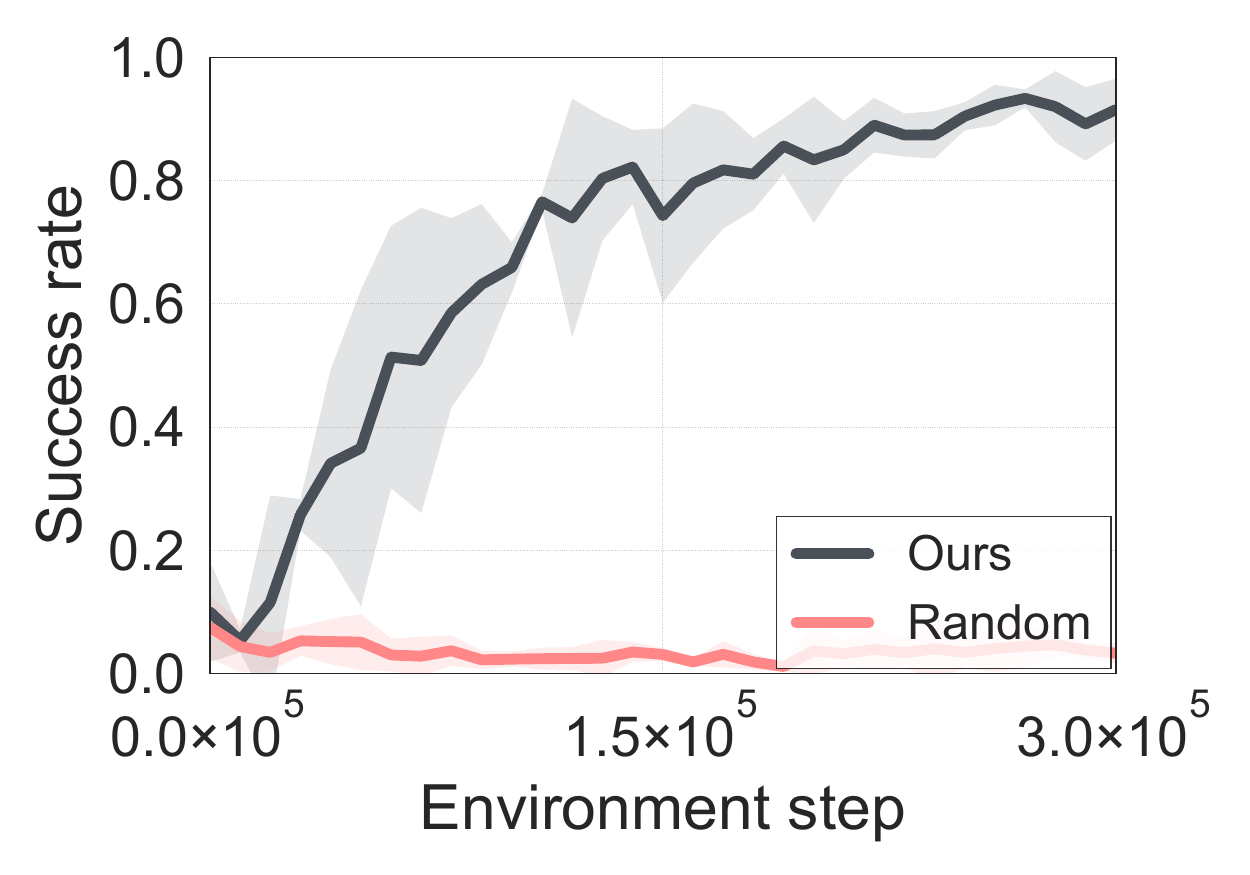}
    \caption{Pusher}
    \label{fig:pusher_random_skip}
    \end{subfigure}
    \caption{Ablation studies about skipping strategy. We compare our skipping strategy to an alternative one: random skipping on (a) 2DReach and (b) Pusher.}
\end{figure}

We compare our subgoal skipping strategy to a simple baseline: random sampling of subgoals from the planned path. As shown in the Figure~\ref{fig:2dreach_random_skip} and \ref{fig:pusher_random_skip}, we find that the alternative performs close to ours in 2DReach, but ours outperforms in Pusher. Developing better skipping strategy is an interesting direction to explore. 

\subsection{Hyperparameter tuning cost.}
\label{supp:tuning_cost}
\begin{figure}[h]
  \begin{subfigure}[b]{0.3\columnwidth}
    \includegraphics[width=\linewidth]{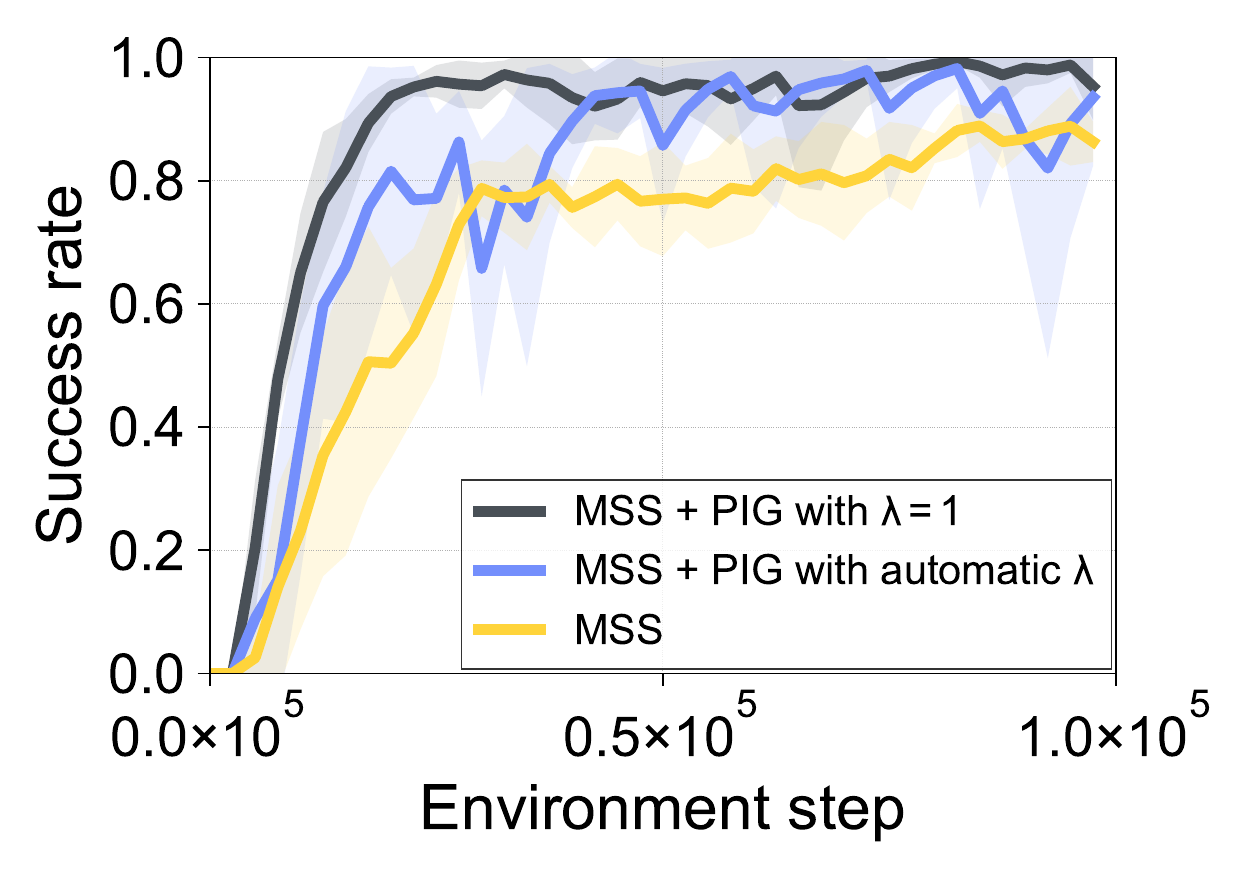}
    \caption{2DReach}
    \label{fig:lambda_2dreach}
  \end{subfigure}
  \hfill 
  \begin{subfigure}[b]{0.35\columnwidth}
  \centering
    \includegraphics[width=0.85714\linewidth]{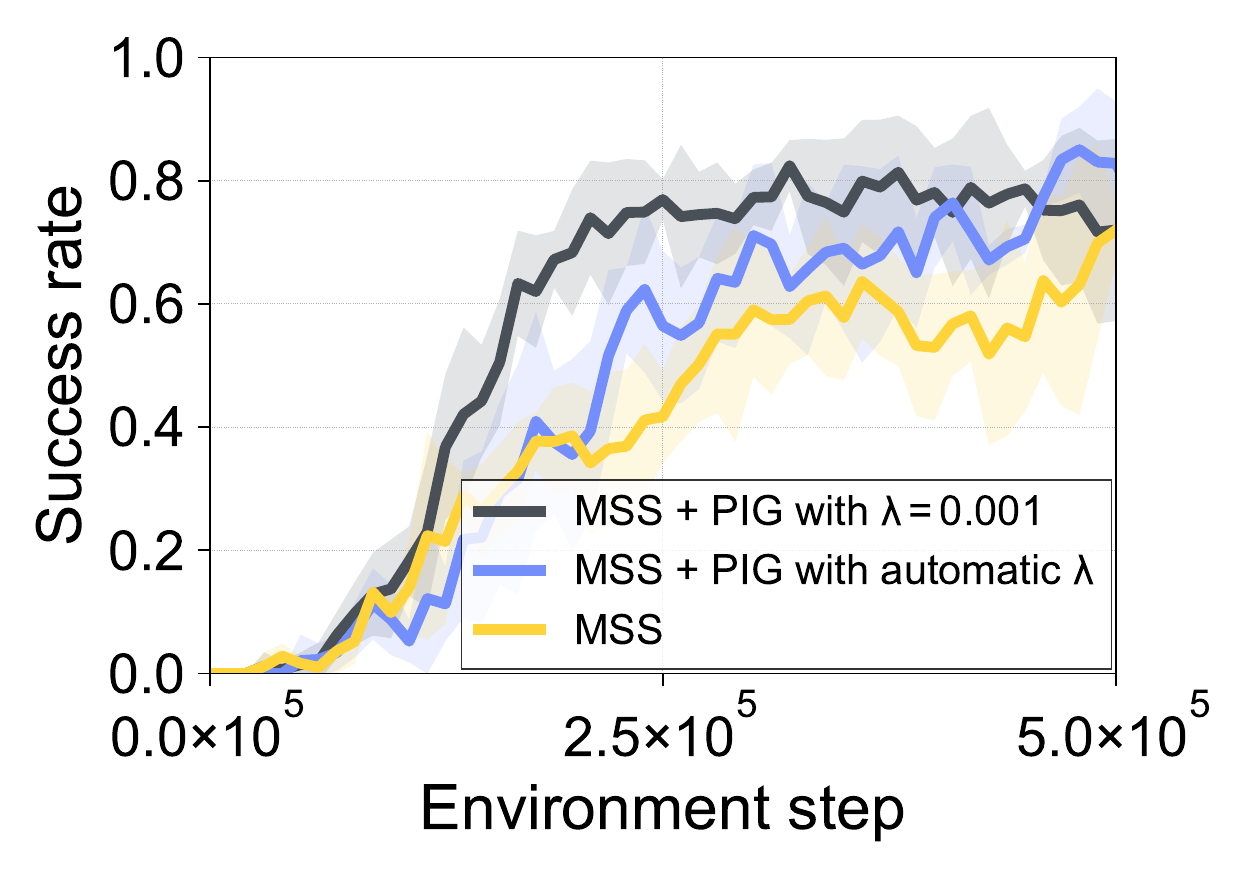}
    \caption{L-shaped AntMaze}
    \label{fig:lambda_antmaze_l}
  \end{subfigure}
  \hfill 
  \begin{subfigure}[b]{0.3\columnwidth}
    \includegraphics[width=\linewidth]{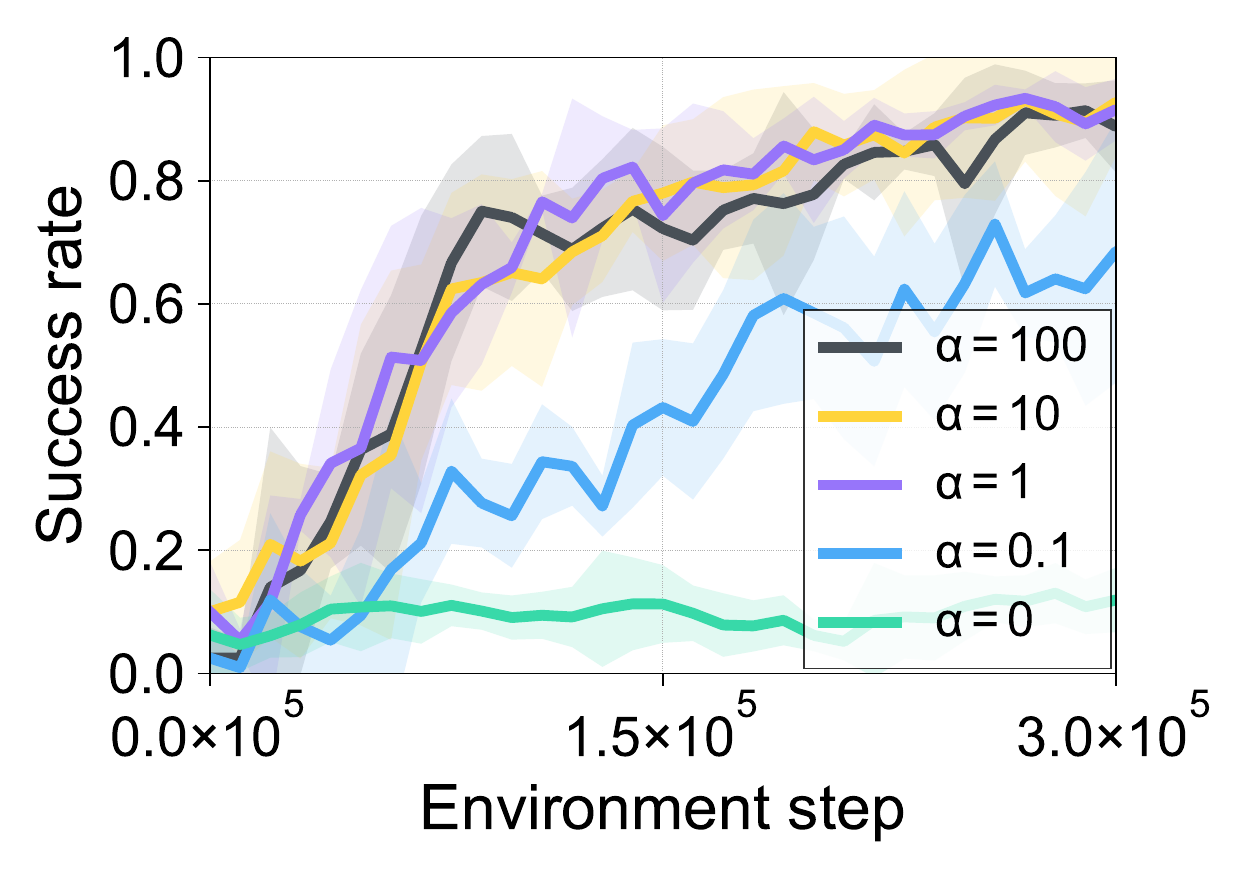}
    \caption{Pusher}
    \label{fig:alpha_pusher}
  \end{subfigure}
  \caption{Experiments with (a, b) automatic hyperparameter setting of $\lambda$ and (c) varying $\alpha$.}
\end{figure}

Our \ALGname inevitably introduces new hyperparameters ($\lambda$ and $\alpha$) in addition to existing algorithms, but we can use a task-agnostic strategy to choose them without any computational overhead. To be specific, one can set the balancing coefficient $\lambda$ adaptively to satisfy $\lambda \times \mathcal{L}_{\mathtt{PIG}} = 0.01 \times \mathcal{L}_{\mathtt{actor}}$; see Figure~\ref{fig:lambda_2dreach}, \ref{fig:lambda_antmaze_l}. Next, we found that the performance of our algorithms is robust to the choice of skipping temperature $\alpha$; see Figure~\ref{fig:alpha_pusher}.

\newpage
\subsection{Effect of subgoal skipping in exploration.}
\label{supp:subgoal_expl}
\begin{wrapfigure}{r}{0.325\linewidth}
\vspace{-.2in}
\centering
    \includegraphics[width=1.0\linewidth]{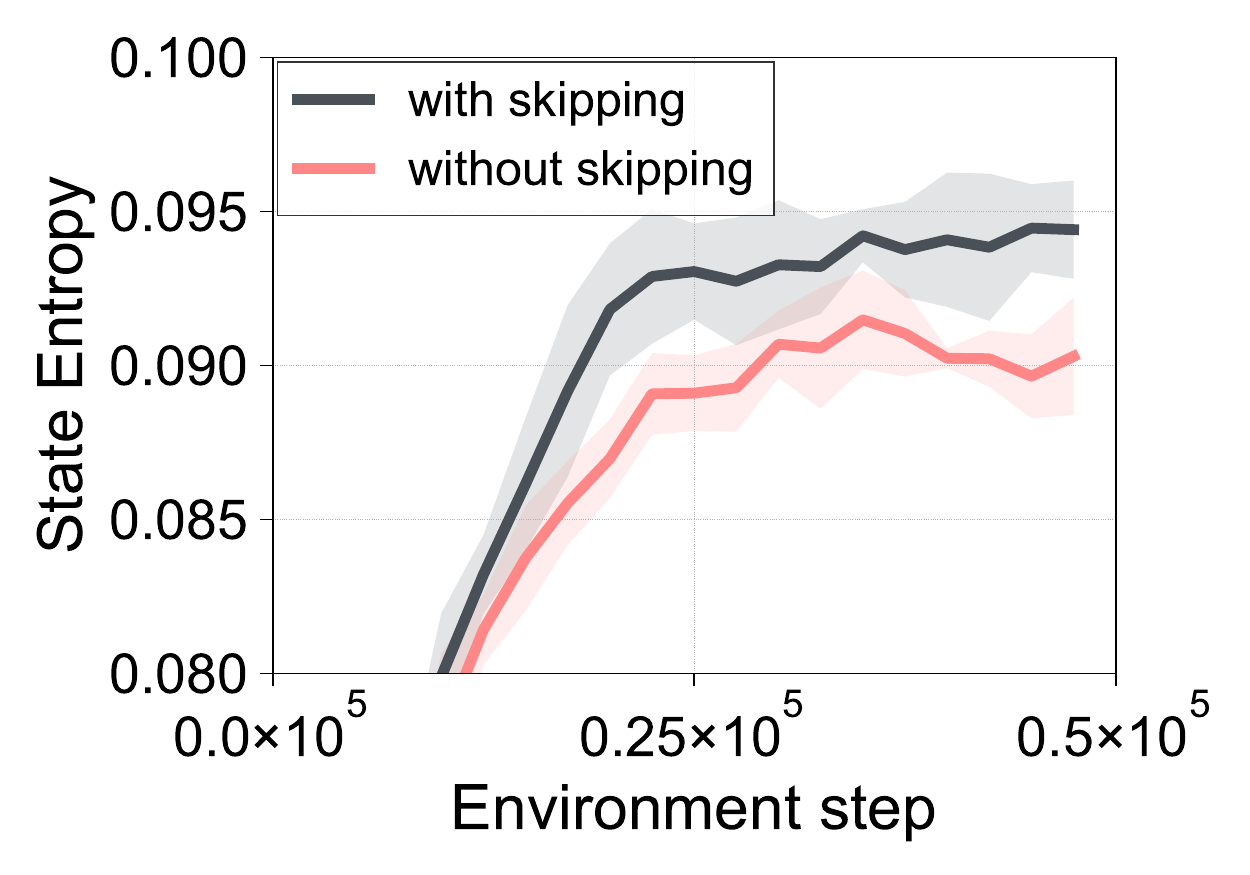}
    \caption{State entropy}
    \label{fig:state_entropy}
\vspace{-.2in}
\end{wrapfigure}
To further support our statement - subgoal skipping makes an agent could collect better trajectories via promoting exploration, we quantitatively measure how diverse an agent discovers states during training depending on subgoal skipping. Specifically, we employ particle-based k-nearest neighbors ($k$-NN) entropy estimator \citep{singh2003nearest} to measure how diverse collected samples are. Formally, let $X$ be a random variable whose probability density function is $p$, and $\{ x_{i} \}_{i=1}^{N}$ be its $N$ i.i.d realization. State entropy is defined as $\mathcal{H} (X) = - \mathbb{E}_{x \sim p(x)} [\log p(x)] $ and we can estimate $\mathcal{H} (X)$ as follows:
\begin{align}
\label{eq:state_entropy}
\hat{\mathcal{H}}_{N}^{K} (X) \propto \frac{1}{N} \sum_{i=1}^{N} \log \frac{1}{K} \sum_{k=1}^{K} \Vert x_{i} - x_{i}^{k - \text{NN} }\Vert_{2},
\end{align}
where $x_{i}^{k-\text{NN}}$ is the $k$-NN of $x_{i}$ within a set $\{ x_{i} \}_{i=1}^{N}$. We use $N=128$ and $K=10$ for an experiment using 2DReach environment.
As shown in Figure~\ref{fig:state_entropy}, we observe that using subgoal skipping makes high state entropy; that is, subgoal skipping makes an agent collect more diverse samples, which is likely to have more chance to include better samples.

\subsection{Reaching a goal without a planner at test time with a larger maze.}
\label{supp:larger_maze_without_planner}
\begin{wrapfigure}{r}{0.325\linewidth}
\vspace{-.2in}
\centering
    \includegraphics[width=1.0\linewidth]{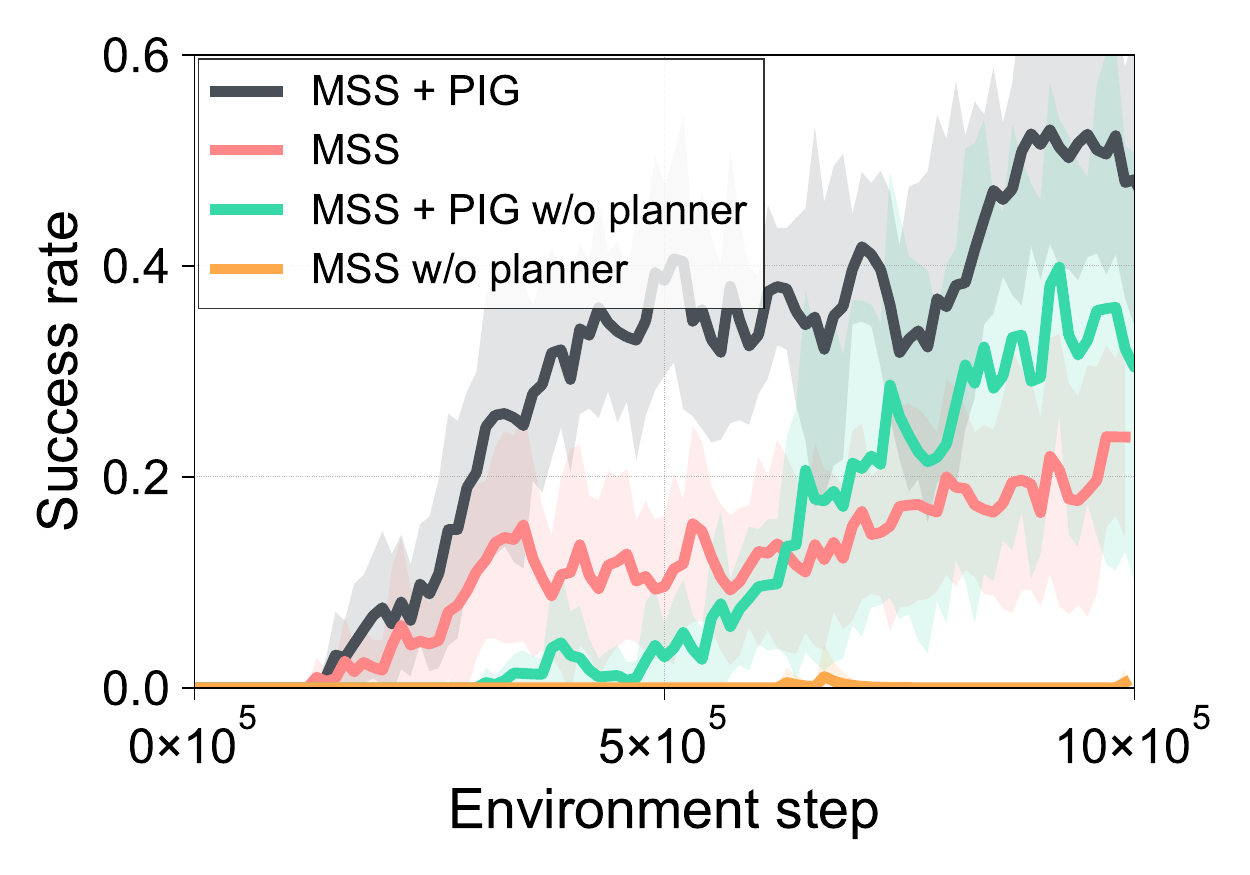}
    \caption{Test time success rate of \ALGname and MSS on large U-shaped Ant Maze over four runs.}
    \label{fig:reaching_without_planner_large}
\vspace{-.8in}
\end{wrapfigure}

We also evaluate without a planner at test time with a large U-shaped AntMaze. As shown in Figure~\ref{fig:reaching_without_planner_large}, training with \ALGname enables successfully reaching the target-goal even without the planner at test time even in larger environment. Intriguingly, after $5 \times 10^5$ environment timesteps, a policy trained by our approach performs better even without access to a planner at test time compared to MSS, which uses a planner at test time. 

\newpage
\subsection{Experiments with stochastic transition model.}

\begin{wrapfigure}{r}{0.325\linewidth}
\vspace{-.2in}
\centering
    \includegraphics[width=1.0\linewidth]{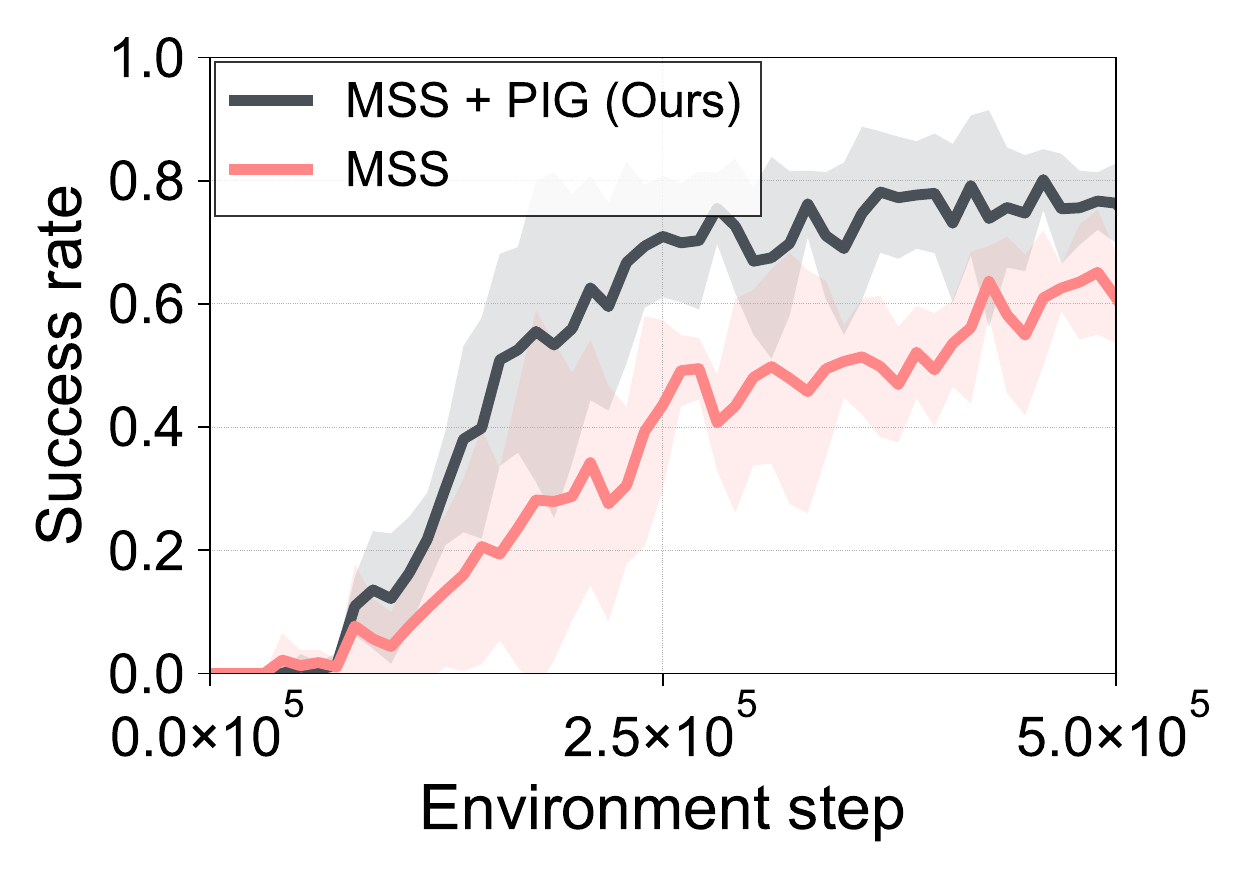}
    \caption{Learning curves on stochastic L-shaped AntMaze as measured on the success rate.}
    \label{fig:stochastic_antmazel}
\vspace{-.2in}
\end{wrapfigure}

\ALGname, along with our graph construction technique, is applicable to stochastic environments since our algorithmic component (self-imitation loss and subgoal skipping) and graph construction mechanism (farthest point sampling and assigning edge weights) are built on visited state spaces, regardless of transition dynamics.

To empirically show that \ALGname is effective in stochastic environments, we additionally provide experimental results on stochastic L-shaped AntMaze, where gaussian noise $\mathcal{N}(0, 0.05)$ is added to the $(x, y)$ position of an agent at every step following setups from \citet{zhang2020generating, kim2021landmark}. As shown in the Figure~\ref{fig:stochastic_antmazel}, we observe that \ALGname successfully solves tasks in the stochastic environment. Moreover, not only in (deterministic) L-shaped AntMaze, but also in stochastic L-shaped AntMaze, PIG shows significant gain compared to the baseline (MSS). This result supports that \ALGname trains a strong policy that is able to reach faraway goals more sample-efficiently than the baseline thanks to our self-imitation loss and subgoal skipping.

\subsection{Ablation studies with more environments.}
\begin{figure*}[h]
    \centering
    \begin{subfigure}{0.325\textwidth}
    \includegraphics[width=1.0\linewidth]{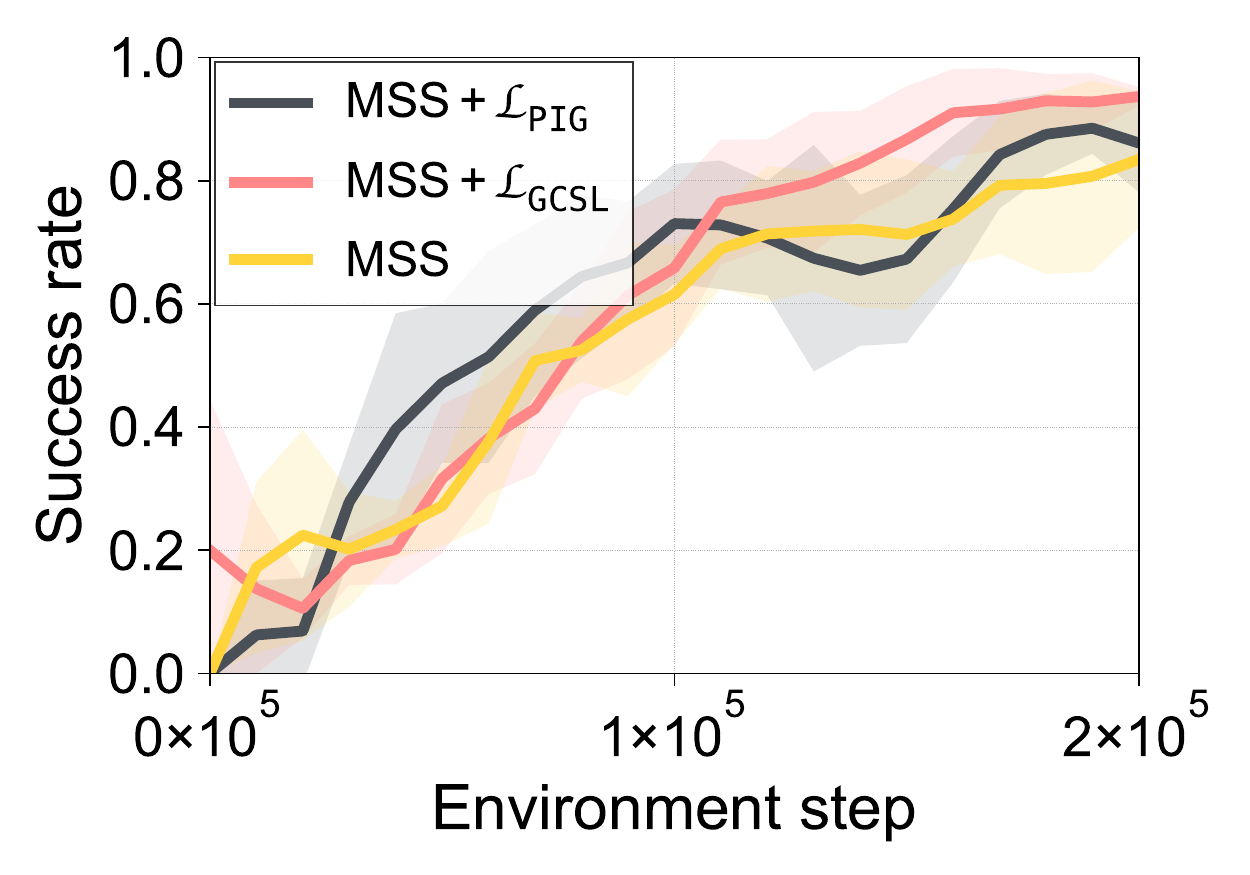}
    \caption{Reacher}
    \end{subfigure}
    \begin{subfigure}{0.325\textwidth}
    \includegraphics[width=1.0\linewidth]{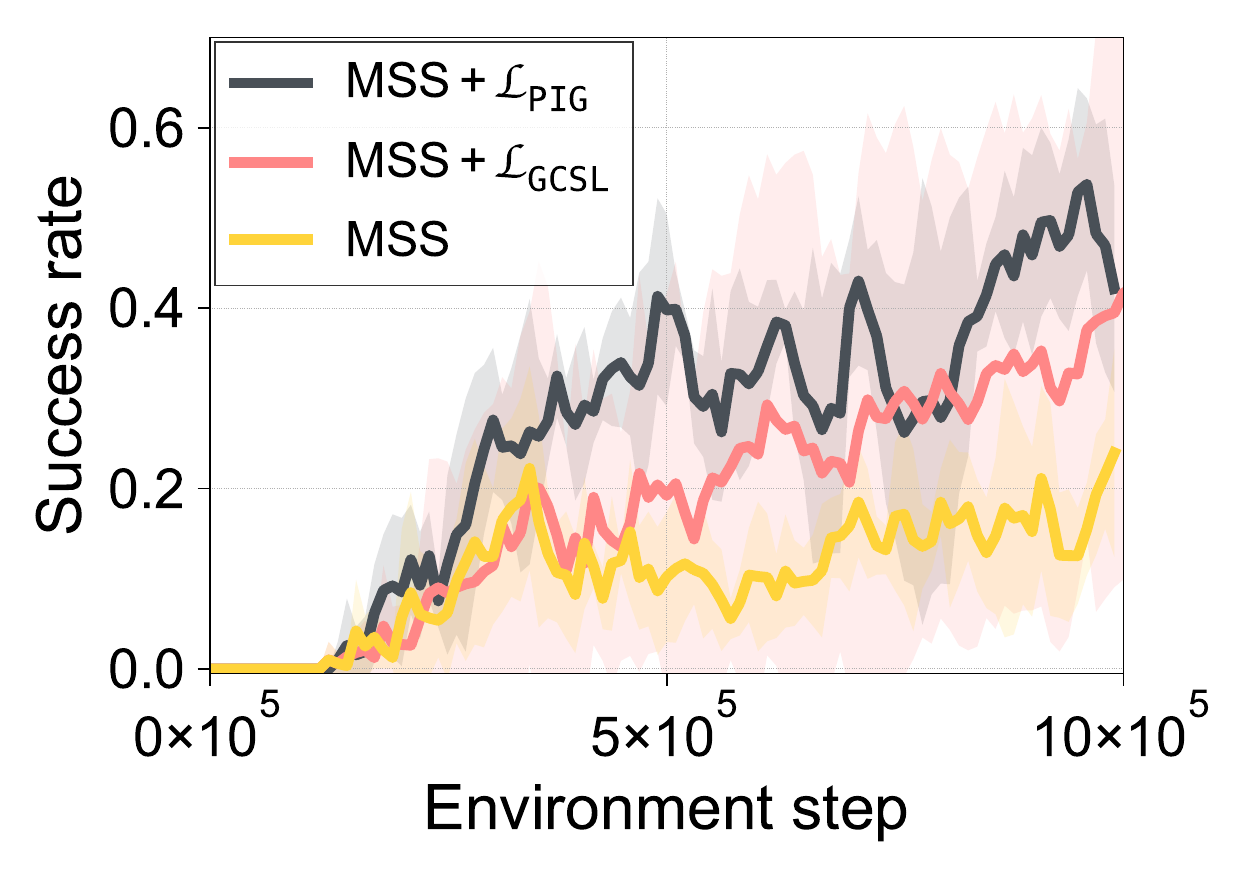}
    \caption{Large U-shaped AntMaze}
    \end{subfigure}
    \caption{Ablation sutides about self-imitation learning for training on (a) Reacher and (b) Large U-shaped AntMaze with four runs. MSS + $\mathcal{L}_{\mathtt{\ALGname}}$ and MSS + $\mathcal{L}_{\mathtt{GCSL}}$ refer to an algorithm that applies loss term $\mathcal{L}_{\mathtt{\ALGname}}$ and $\mathcal{L}_{\mathtt{GCSL}}$ on top of MSS method, respectively; subgoal skipping is not applied. We find that our loss term $\mathcal{L}_{\mathtt{\ALGname}}$ is more effective than $\mathcal{L}_{\mathtt{GCSL}}$ as an auxiliary term.}
    \label{supp:abaltion_loss}
\end{figure*}

\begin{figure*}[h]
    \centering
    \begin{subfigure}{0.325\textwidth}
    \includegraphics[width=1.0\linewidth]{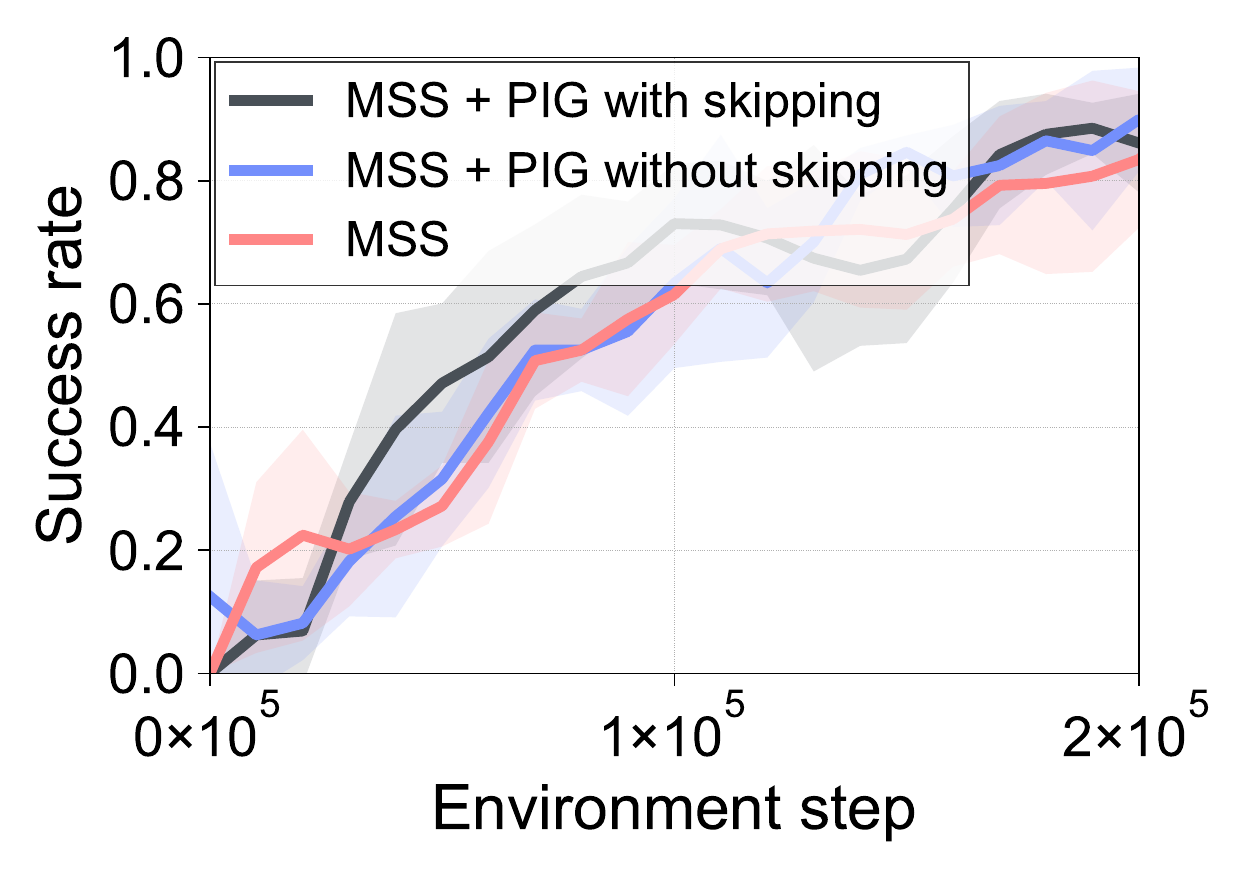}
    \caption{Reacher}
    \end{subfigure}
    \begin{subfigure}{0.325\textwidth}
    \includegraphics[width=1.0\linewidth]{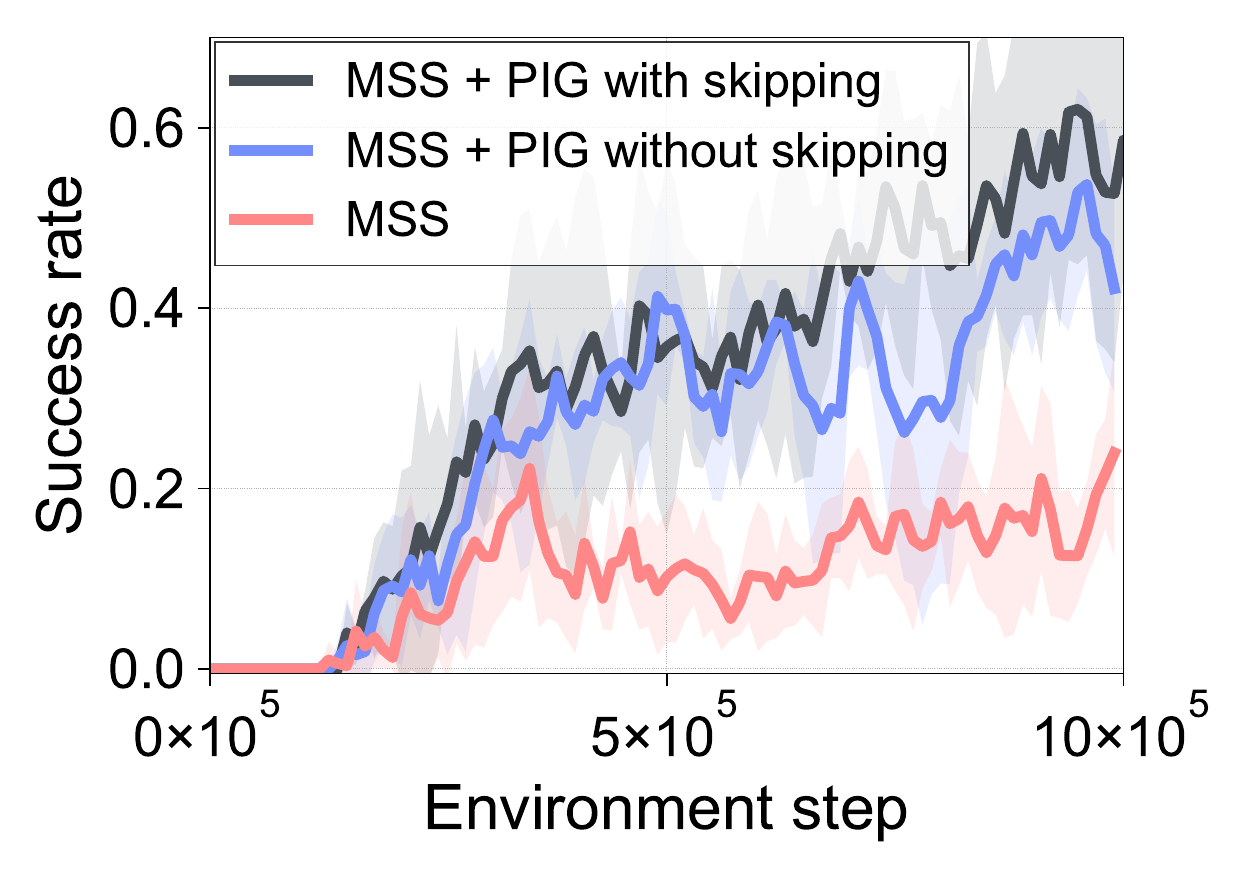}
    \caption{Large U-shaped AntMaze}
    \end{subfigure}
    \caption{Learning curves of PIG with and without subgoal skipping on (a) Reacher and (b) Large U-shaped AntMaze tasks with four runs.}
    \label{supp:ablation_skipping}
\end{figure*}

We provide ablation studies about self-imitation loss and subgoal skipping with more environments: Reacher and Large U-shaped AntMaze. As showin in Figure~\ref{supp:abaltion_loss} and \ref{supp:ablation_skipping}, including our self-imitation loss or subgoal skipping makes significant gains or performs on par.

\newpage

\subsection{Experiments with extended timesteps.}
\label{supp:extended_timesteps}
\begin{wrapfigure}{r}{0.325\linewidth}
\vspace{-.2in}
\centering
    \includegraphics[width=1.0\linewidth]{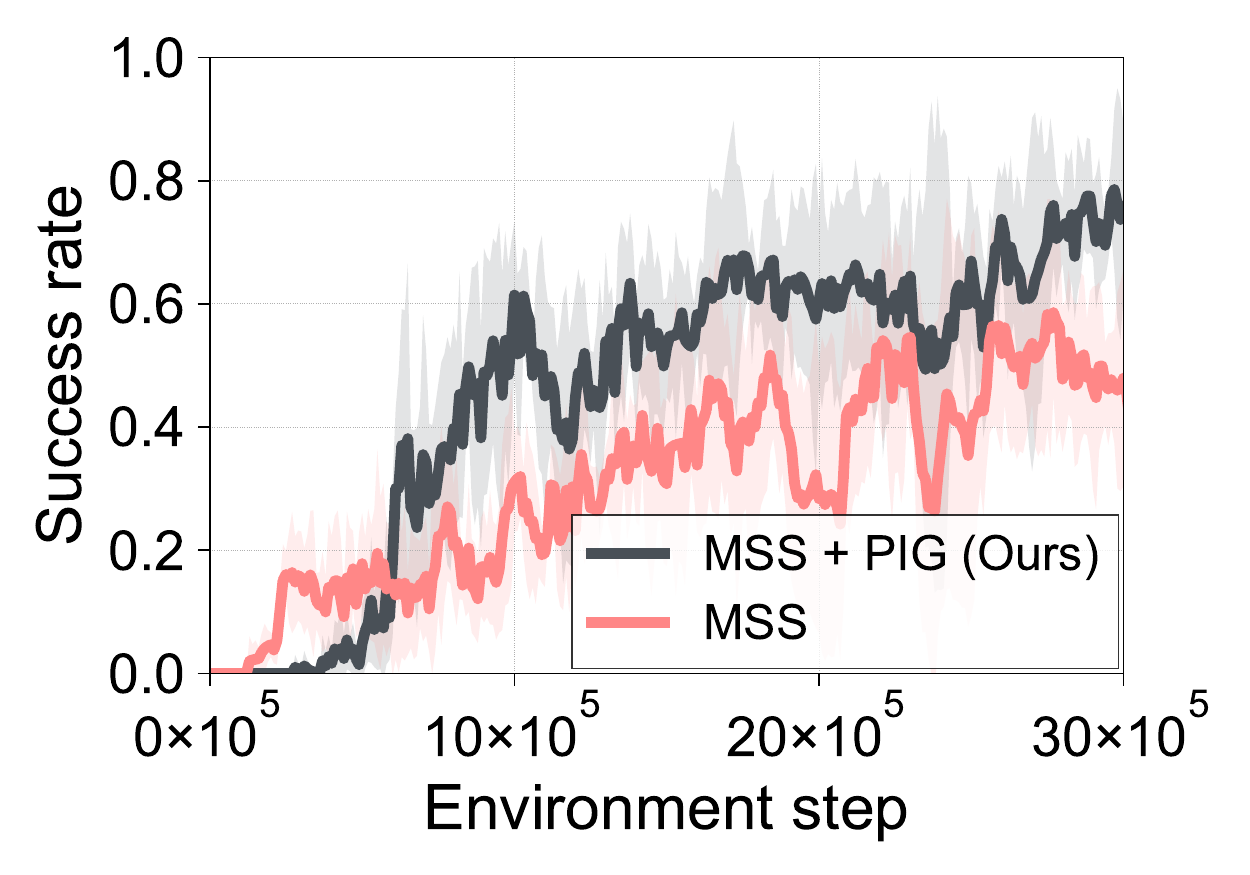}
    \caption{Learning curves on U-shaped AntMaze as measured on the success rate.}
    \label{fig:extended}
\vspace{-1in}
\end{wrapfigure}

To assess whether the empirical improvements are in learning speed or also in asymptotic performance,
we evaluate PIG and MSS with extended timesteps (i.e., from $10 \times 10^5$ to $30 \times 10^5$ on Large U-shaped AntMaze. 
As shown in the Table below, we find that PIG can improve both sample-efficiency and asymptotic performances of MSS. This shows that enhanced policy learning via information distillation from the planner can also improve the asymptotic performance.

\vspace{.8in}

\section{Implementation details}
\label{supp:impl}
All of the experiments were processed using a single GPU (NVIDIA TITAN Xp) and 8 CPU cores (Intel Xeon E5-2630 v4).
For baselines, we employ open-source codes of MSS\footnote{\url{https://github.com/FangchenLiu/map\_planner}}, \LthreeP\footnote{\url{https://github.com/LunjunZhang/world-model-as-a-graph}}, and HIGL\footnote{\url{https://github.com/junsu-kim97/HIGL}}.
\subsection{Graph construction}
\label{supp:graph_construction}
\textbf{Collection of graph-constructing states.} 
We follow collecting scheme of graph-construction states from \citet{huang2019mapping}. The collection is proceeded in two steps: (a) random sampling of a fixed-sized pool $\mathcal{D}$ from an experience replay and (b) farthest point sampling (FPS) \citep{vassilvitskii2006k, huang2019mapping} from the pool $\mathcal{D}$ to build the final collection $\mathcal{V}$ of graph-constructing states. 

Specifically, any given time, let $D(s)$ denote the shortest distance from a state $s$ to the closest element in current $\mathcal{V}$. The set $\mathcal{V}$ is initialized with an empty set.
Then, FPS runs as follows:
\begin{itemize}[topsep=1.0pt,itemsep=1.0pt,leftmargin=5.5mm]
    \item Step A: Choose a state $s^{1}$ uniformly at random from the pool $\mathcal{D}$ and add $s^{1}$ into $\mathcal{V}$.
    \item Step B: Choose the next state $s^{i}$, whose $D(s^{i})$ is the largest among elements in $\mathcal{D}$. Add $s^{i}$ into $\mathcal{V}$.
    \item Step C: Repeat Step B until we have chosen a budget for the number of nodes in a graph. 
\end{itemize}

The diversity of the collection is ensured by farthest point sampling. Random sampling to build a fixed-sized pool makes the computational complexity of planning irrelevant to the size of experience replay, of which size is 1M in our experiments.

\textbf{Edge connection.}
After collecting graph-constructing states, we complete a graph by adding directed edges \citep{huang2019mapping}. 
In detail, given two nodes $l^{1}$ and $l^{2}$, we connect them by adding two directed edges $(l^{1}, l^{2}) \in \mathcal{E}$ (from $l^{1}$ to $l^{2}$) and $(l^{2}, l^{1}) \in \mathcal{E}$ (from $l^{2}$ to $l^{1}$). Then we assign weights as an estimated distance $d(l^{1}, l^{2})$ and $d(l^{2}, l^{1})$, respectively.

\subsection{Hyperparameters}
\label{supp:hyperparameters}
We list hyperparameters used for \ALGname across all environments in Table~\ref{tbl:hyperparameters_common} and \ref{tbl:hyperparameters_specific}. 

For the baselines, we used the best hyperparameters reported in their source codes for shared environments: 2DReach of MSS and HER, Reacher and Pusher for HIGL, and AntMazes for MSS, L3P, HER, and HIGL (all). For unstudied environments in the baseline papers, we have searched hyperparameters for each baseline. For example, we search shift magnitude and adjacency degree for HIGL, clipping threshold and final goal adjacency threshold for L3P and MSS, and relabeling ratio for HER. We note that for PIG, two newly introduced hyperparameters (balancing coefficient $\lambda$ and skipping temperature $\alpha$) have been searched. We would like to remark that performance gain by PIG have been achieved without exhaustive efforts in hyperparameter search compared to baselines. For example, the baseline MSS conducted grid search on 30 (number of landmarks) $\times$ 30 (clipping threshold) values in their paper, but we searched among 5 $\times$ 4 values for PIG: $\{1.0, 0.1, 0.01, 0.001, 0.0001\}$ for $\lambda$ and $\{20, 10, 5, 1\}$ for $\alpha$.

\begin{table}[h]
\caption{Hyperparameters across all environments.}
\vskip 0.15in
\centering
\begin{center}
\begin{tabular}{l c}
\toprule

\textbf{Hyperparameter} & \textbf{Value} \\
\midrule
\textit{DDPG} & \\
\midrule
Optimizer & Adam
\citep{kingma2014adam} \\
Actor learning rate & 0.0002 \\
Critic learning rate & 0.0002 \\
Replay buffer size & 1M \\
Number of hidden layers for actors & 4 \\
Number of hidden layers for critics & 5 \\
Number of hidden units per layer & 400 \\
Batch size & 200 \\
Nonlinearity & ReLU \\
Polyak for target network & 0.99 \\
Target update frequency per episode & 3 \\
Ratio between env vs optimization steps & 1 \\
Gamma & 0.99 \\
Hindsight relabelling ratio & 0.8 \\
\midrule
\midrule
\textit{Graph} & \\
\midrule
Number of soft value iteration & 20 \\
Temperature & 0.9 \\
\bottomrule
\end{tabular}
\label{tbl:hyperparameters_common}
\end{center}
\vskip -0.1in
\end{table} 

\begin{table}[h]
\caption{Hyperparameters that differ across the environments.}
\vskip 0.15in
\centering
\large
\begin{center}
\resizebox{\textwidth}{!}{
\begin{tabular}{l cccc}
\toprule

\textbf{Hyperparameter} & \textbf{2DReach} & \textbf{Reacher} & \textbf{Pusher} & \textbf{AntMaze} \\
\midrule
\textit{Ours-specific} & \\
\midrule
Balancing coefficient $\lambda$        & 1.0 & 0.0001 & 0.1 & 0.001 \\
Skipping temperature $\alpha$          & 1.0 & 10.0   & 1.0 & 10.0  \\
\midrule
\midrule
\textit{DDPG} & \\
\midrule
Initial random trajectories    & 2.5k & 20k & 20k & 100k (for L-, U- shaped Maze) \\
                               &      &     &     & 400k (for Large U-shaped Maze) \\   
                               &      &     &     & 800k (for S-, $\omega$-, $\Pi$ -shaped Maze) \\   
Hindsight relabelling range & 50 & 50 & 50 & 200 \\

Action L2 & 0.5 & 0.01 & 0.01 & 0.5 \\
Action noise & 0.2 & 0.1 & 0.1 & 0.2 \\
\midrule
\midrule
\textit{Graph} & \\
\midrule
Number of nodes in a graph & 100 & 80 & 80 & 400 \\
clipping threshold for distances & 4.0 & 4.0 & 4.0 & 38.0 \\

\bottomrule
\end{tabular}
\label{tbl:hyperparameters_specific}
}
\end{center}
\vskip -0.1in
\end{table} 


\end{document}